\documentclass{article}

\usepackage[final]{neurips_2021}

\usepackage[utf8]{inputenc} 
\usepackage[T1]{fontenc}    
\usepackage{hyperref}       
\usepackage{url}            
\usepackage{booktabs}       
\usepackage{amsfonts}       
\usepackage{nicefrac}       
\usepackage{microtype}      
\usepackage{xcolor}         
\usepackage{amsmath}
\usepackage{algorithm}
\usepackage{algorithmic}
\usepackage{ulem}
\usepackage{enumitem}
\usepackage{graphicx}
\usepackage{bm}
\usepackage{multirow}
\usepackage{bbm}
\usepackage{subfigure}
\usepackage{caption}

\title{AutoGEL: An Automated Graph Neural Network \\
with Explicit Link Information
}

\author{
  Zhili WANG, \ Shimin DI\thanks{Corresponding author}\ 
  , \ Lei CHEN \\
  Department of Computer Science and Engineering\\
  The Hong Kong University of Science and Technology\\
  \texttt{ \{zwangeo,sdiaa,leichen\}@connect.ust.hk}  
}

\begin{document}

\maketitle

\begin{abstract}
Recently, Graph Neural Networks (GNNs) have gained popularity in a variety of real-world scenarios. Despite the great success, the architecture design of GNNs heavily relies on manual labor. Thus, automated graph neural network (AutoGNN) has attracted interest and attention from the research community, which makes significant performance improvements in recent years. However, existing AutoGNN works mainly adopt an implicit way to model and leverage the link information in the graphs, which is not well regularized to the link prediction task on graphs, and limits the performance of AutoGNN for other graph tasks. In this paper, we present a novel AutoGNN work that explicitly models the link information, abbreviated to AutoGEL. In such a way, AutoGEL can handle the link prediction task and improve the performance of AutoGNNs on the node classification and graph classification task. Specifically, AutoGEL proposes a novel search space containing various design dimensions at both intra-layer and inter-layer designs and adopts a more robust differentiable search algorithm to further improve efficiency and effectiveness. Experimental results on benchmark data sets demonstrate the superiority of AutoGEL on several tasks.
\end{abstract}

\section{Introduction}
\label{sec:intro}

As one of ubiquitous data structures,
graph $G(E,V)$
contains the node-set $V=\{v_1,\cdots,v_n\}$
and 
edge-set $E=\{e(v_i,v_j): v_i,v_j\in V\}$,
which can represent a lot of real-world data sets, such as 
social networks \citep{bu2018gleam}, physical systems \citep{sanchez2018graph}, protein-protein interaction networks \citep{yue2020graph}.
In recent years, 
Graph Neural Networks (GNNs) have been introduced for 
various graph tasks and achieve unprecedented success,
such as node classification \citep{kipf2016semi,hamilton2017inductive},
link prediction \citep{vashishth2019composition,li2020distance},
and graph classification \citep{niepert2016learning,zhang2018end}.
Generally,
GNNs encode $G(V,E)$ into the $d$-dimensional 
vector space (e.g., $\mathbf{V}\in\mathbb{R}^{|V|\times d}$ ) that preserves similarity in the original graph. 
Despite the great success, 
those GNNs are restricted to specific instances within GNN design space \citep{you2020design}.
Different graph tasks usually require different GNN architectures \citep{gilmer2017neural}. 
For example, 
compared with the node classification task,
GNNs for graph classification introduces an extra readout phase to obtain graph embeddings.
However,
architecture design for these GNNs remains a challenging problem due to the diversity of graph data sets.
Given a graph task,
a GNN architecture performs well on one 
data set
does not
necessarily imply that it is also suitable for other data sets
\citep{you2020design}.

Some pioneer works have been proposed to alleviate the above issue in GNN models.
They introduce
Neural Architecture Search (NAS)
\citep{elsken2019neural}
approaches to automatically design suitable GNN architecture for the given data set (i.e., AutoGNN) 
\citep{zhou2019auto,gao2020graph,jiang2020graph,zhao2021search}.
Architectures identified by these AutoGNNs rival or surpass the best handcrafted GNN models, demonstrating the potential of AutoGNN  towards better GNN architecture design.
Unfortunately,
the 
existing AutoGNNs 
are mainly designed for the node classification and graph classification task.
Their designs do not include edge embeddings,
i.e., modeling and organizing link information in an implicit way.
First,
it is difficult for existing AutoGNNs to handle another important graph task on the edge-level,
link prediction (LP) task.
Second,
lack of
edge embedding
makes them 
inexpressive
to leverage the complex link information, such as 
direction information of edges and
different edge types in multi-relational graphs.
Especially,
various edge types could impose different influence for encoding nodes into embeddings,
which can further improve the model performance on the node-level and graph-level tasks.
Therefore, a 
new AutoGNN is desired to model link information explicitly on various data sets.

To bridge the aforementioned research gap, we propose AutoGEL, a novel \textit{AutoG}NN framework
with \textit{E}xplicit \textit{L}ink information,
which can handle the LP task and improve performance of AutoGNNs on other graph tasks.
Specifically,
AutoGEL
explicitly learns the edge embedding in the
message passing framework to model the complex link information,
and
introduces
the several novel design dimensions into the GNN search space,
enabling a more powerful GNN to be searched for any given graph data set.
Moreover, AutoGEL adopts a robust differentiable search algorithm to guarantee the effectiveness of searched architectures and 
control the computational footprint.
We summarize the contributions of this work as follows:
\begin{itemize}[leftmargin=*]

\item 
The design of existing AutoGNNs 
follows an implicit way to 
leverage and 
organize the link information, which cannot
handle the LP task and limits the performance
of AutoGNNs on other graph tasks.
In this paper, we present a novel method called AutoGEL
to solve these issues through explicitly modeling 
the link information in the graphs.

\item 
AutoGEL introduces several novel design dimensions
into the GNN search space at both the intra-layer and inter-layer designs, 
so as to improve the task performance.
Moreover,
motivated by one robust NAS algorithm SNAS,
AutoGEL upgrades the search algorithm adopted in existing AutoGNNs to further 
guarantee the effectiveness of final derived GNN.


\item 
The experimental results demonstrate 
that 
AutoGEL can achieve better performance than manually designed models in the LP task.
Furthermore,
AutoGEL 
shows excellent competitiveness with other AutoGNN works on the node and graph classification tasks.

\end{itemize}

\section{Related Work}

\subsection{General Message Passing Framework}

The majority of GNNs follow the neighborhood aggregation schema \citep{gilmer2017neural}, i.e., the Message Passing Neural Network (MPNN), which is formulated as:
\begin{align}
\label{eq:mp1}
& \mathbf{m}_{v}^{k+1} = AGG_k(\{M_k(\mathbf{h}^k_v, 
\mathbf{h}^k_u, \mathbf{e}_{vu}^k): u\in N(v)\}), \ \mathbf{h}_{v}^{k+1} = ACT_k(COM_k(\{\mathbf{h}_{v}^k,\mathbf{m}_{v}^{k+1}\})), \\
\label{eq:mp2}
& \hat{\mathbf{y}} = R(\{\mathbf{h}^L_v | v\in G\}),
\end{align}
where $k$ denotes $k$-th layer, 
$N(v)$ denotes a set of neighboring nodes of $v$, $\mathbf{h}^k_v$, $\mathbf{h}^k_u$ denotes hidden embeddings for $v$ and $u$ respectively, $\mathbf{e}_{vu}^k$ denotes features for edge $e(v,u)$ (optional), 
$\mathbf{m}_{v}^{k+1}$ denotes the intermediate embeddings gathered from neighborhood $N(v)$,
$M_k$ denotes the message function,
$AGG_k$ denotes the neighborhood aggregation function, 
$COM_k$ denotes the combination function between intermediate embeddings and embeddings of node $v$ itself from the last layer, 
$ACT_k$ denotes activation function.
Such message passing phase in \eqref{eq:mp1} repeats for $L$ times (i.e., $k\in\{1,\cdots,L\}$). 
For graph-level tasks, it further follows the readout phase in \eqref{eq:mp2} where information from the entire graph $G$ is aggregated through readout function $R(\cdot)$.


\begin{table}[t]
	\caption{Overview of Existing AutoGNN Works. 
		The ``Differ.'' denotes to differentiable algorithm.
	}
	\label{tab:autognn}
	\setlength\tabcolsep{1pt}
	\centering
	\begin{tabular}{c|c|c|c|c|c|c}
		\toprule
		\multirow{2}{*}{\bf Method} & \bf Graph &  
		\multicolumn{3}{c|}{\bf MPNN Space}
		&\multirow{2}{*}{\bf Search Algorithm}& \multirow{2}{*}{\bf Task} \\ \cmidrule{2-5}
		&  \#node/edge types & $\mathbf{h}_{e}$ &  intra & inter && \\
		\midrule
		GraphNAS & $\geq 1 / =1$ & $\times$ & $\surd$ & $\times$  &RL&node \\
		AGNN & $\geq 1 / =1$ & $\times$ & $\surd$ & $\times$ &EA+RL &node \\
		SANE & $\geq 1 / \geq 1$ & $\times$ & $\surd$ &$layer\_cnt\&\_agg$ & deterministic Differ. &node \\
		NAS-GCN & $\geq 1 / \geq 1$ & $\times$ & $\surd$&$layer\_cnt$ &EA & graph \\
		\cite{you2020design}& $\geq 1 / =1$ & $\times$ &$\surd$&$layer\_cnt$& random &node/edge/graph \\
		\midrule
		AutoGEL & $\geq 1 / \geq 1$ & $\surd$ & $\surd$ &$layer\_cnt\&\_agg$& stochastic Differ. & node/edge/graph \\
		\bottomrule
	\end{tabular}
	\vspace{-15px}
\end{table}

\subsection{Automated Graph Neural Networks (AutoGNN)}
\label{ssec:autognn}

In recent years, AutoGNN has emerged as a promising direction towards better graph neural architecture design \citep{zhou2019auto,gao2020graph,jiang2020graph,you2020design,zhao2021search,ding2021diffmg}.
To enable a powerful GNN architecture to be searched,
AutoGNNs first propose the GNN search space in the \textit{intra-layer} level, i.e., providing common choices for several important operators in one MPNN layer (see \eqref{eq:mp1} and \eqref{eq:mp2}).
Here we summarize candidate choices for those operators:

\begin{itemize}[leftmargin=*]
	
	\item \textbf{Message Function} $M_k$:
	Existing AutoGNNs mainly
	focus on the node-level and graph-level task, thus 
	edge embeddings are often not available.
	$M_k(\mathbf{h}_v, \mathbf{h}_u,\mathbf{e}_{vu})$ is reduced to $M_k(\mathbf{h}_v, \mathbf{h}_u)$. 
	Typically, $M(\mathbf{h}_v, \mathbf{h}_u) = a_{vu}\mathbf{W}\mathbf{h}_u$, where $a_{vu}$ denotes the attention scores, and 
	$\mathbf{W}\in \mathbb{R}^{d\times d}$
	denotes the linear transformation matrix.
	Note that NAS-GCN \citep{jiang2020graph}
	only takes the edge feature $\mathbf{e}_{vu}$ as input without learning edge embeddings.
	We next denote the edge embedding to $\mathbf{h}_{e}$ for distinguishing.
	
	\item \textbf{Aggregation} $AGG_k$: 
	{It controls the way to aggregate message from nodes' neighborhood.}
	It can be any differentiable and permutation invariant functions, usually
	$AGG_k \in [sum, mean, max]$.
	And 
	$sum(\cdot) = \sum_{u \in N(v)}M_k(\mathbf{h}_v, \mathbf{h}_u)$, $mean(\cdot) = \sum_{u \in N(v)}M_k(\mathbf{h}_v, \mathbf{h}_u)/|N(v)|$, and 
	$max(\cdot)$ denotes channel-wise maximum across the node dimension.

	\item \textbf{Combination} $COM_k$: 
	It determines the way to merge messages between neighborhood and node itself.
	In literature, $COM_k$ is selected from $[concat, add, mlp]$,
	where $concat(\cdot)=[\mathbf{h}^k_v,\mathbf{m}^{k+1}_v]$, 
	$add(\cdot)=\mathbf{h}^k_v+\mathbf{m}^{k+1}_v$,
	and $mlp(\cdot)=\mathbf{MLP}(\mathbf{h}^k_v+\mathbf{m}^{k+1}_v)$ ($\mathbf{MLP}$ is Multi-layer Perceptron).

	\item \textbf{Activation} $ACT_k$: $[identity, sigmoid, tanh, relu, elu]$ are some of the most commonly used activation functions \citep{gao2020graph}.
	
	\item \textbf{Graph Pooling:} The pooling operator has been introduced for the graph classification task, such as $[global\ pool, global\ attention\ pool, flatten]$ \citep{jiang2020graph,wei2021pooling}.

\end{itemize}

In addition to the above intra-layer operators, several works \citep{you2020design,zhao2021search,jiang2020graph}
propose the idea of searching layer connectivity to combine hidden representations of different layers in a better way, i.e.,
\textit{inter-layer} design. 
More details will be discussed in Sec.~\ref{sssec:inter}.
After the search space design,
AutoGNNs adopt various search algorithms to search the optimal architecture from the search space.
AGNN \citep{zhou2019auto}
and 
GraphNAS \citep{gao2020graph} 
follow the reinforcement learning (RL)
\citep{williams1992simple}
way to search architectures.
They
utilize a recurrent neural network controller for architecture sampling, and update the controller to maximize the expected performance of sampled architectures.
NAS-GCN \citep{jiang2020graph} adopt evolutionary algorithm (EA), where new architectures are generated by performing mutation from parent architectures and the population, i.e., the best performing architectures, are iteratively updated. 
However,
both RL and EA algorithms
require a large number of architectures to be sampled, which is inherently computational expensive.
To improve the search efficiency,
SANE \citep{zhao2021search} 
adopts a deterministic differentiable search algorithm DARTS \citep{liu2018darts},
where a supernet containing all candidate operators is constructed and architecture parameters are jointly optimized with network parameters through gradient descent.
Unfortunately,
it has been discussed in 
SNAS \citep{xie2018snas} 
that 
DARTS suffer from the unbounded bias issue towards its objective, which limits the performance of the final derived model.
In Tab.~\ref{tab:autognn},
we summarize existing AutoGNNs from several perspectives: 
graph,
MPNN space, search algorithm, and task scenario.

\subsection{GNNs for Link Prediction Task}

Even with the great effort invested into the construction and maintenance of networks, many graph data sets still remain incomplete \citep{schlichtkrull2018modeling}. 
Therefore, the link prediction (LP) task is one of the most crucial problems in the graph-structured data,
which aims to recover
those missing links in a graph~\citep{zhang2019nscaching,zhang2020interstellar},
i.e.,
predicting the missing node in $e(v_i,?)$ where $?$ denotes the target node that has the potential link with $v_i$.
Recently,
GNN models have been introduced to handle the LP task (abbreviated to GLP models),
which can be roughly categorized based on its application scenarios: GLP on homogeneous graphs (i.e., only one type of nodes and edges \citep{yang2016revisiting}),
and multi-relational graphs (i.e., graphs with multiple edge types \citep{toutanova2015observed}).

\subsubsection{Link Prediction on Homogeneous Graphs}


As one of the classic approaches for LP task on such homogeneous graphs,
heuristic methods
predict link existence according to heuristic node similarity scores \citep{zhang2020revisiting}.
Despite its effectiveness in some scenarios, heuristic methods hold strong assumptions on the link formation mechanism, i.e., highly similar nodes have links. 
It would fail on those networks where their assumptions do not hold \citep{zhang2018link}.
Furthermore,
latent feature-based methods \citep{perozzi2014deepwalk,grover2016node2vec}
factorize some network matrices to learn node embeddings in a transductive way, which limits their generalization ability to unseen data.

Recently, several GLP models have been proposed for the LP task on homogeneous graphs. GAE \citep{kipf2016variational} applies GNN model over the entire graph and aims to learn node embeddings that minimize the graph reconstruction cost through:
\begin{align}
\label{eq:gae}
    \mathbf{H}=GCN(\mathbf{X},\mathbf{A}),\hat{\mathbf{A}}=\sigma(\mathbf{H}\mathbf{H}^{\top})
\end{align}
where $\mathbf{X}\in \mathbb{R}^{|V|\times{D}}$ is the feature matrix of nodes, $\mathbf{A}\in \mathbb{R}^{|V|\times{|V|}}$ is the adjacency matrix,
$\mathbf{H}$ is the learned node representations,
$\sigma(\cdot)$ is the logistic sigmoid function,
$\hat{\mathbf{A}}$ denotes the reconstructed adjacency matrix, whose entry $\hat{\mathbf{A}}_{uv}$ is the predicted score (between 0-1) for target link $e_{uv}$.
However, GAE focus on aggregating node attributes only. 

SEAL \citep{zhang2018link} and DE-GNN \citep{li2020distance}
propose to learn the link embedding from the subgraph structures.
Specifically, DE-GNN \citep{li2020distance} considers
feature of subgraph structure
$\mathbf{X}^{sub}$
for aggregation
and utilize node-set level readout:
\begin{equation}
\label{eq:degnn}
\mathbf{H^{sub}}=GCN(\mathbf{X}^{sub},\mathbf{A}^{sub}), 
\
\mathbf{h}_{e} = 
R
(\{\mathbf{h}_v|v \in S\}),
\end{equation}
where $S=\{u,v\}$ denotes two nodes in the link $e(u,v)$ for LP task,
and $R(\cdot)$ is difference-pooling in DE-GNN, i.e., $R(\cdot)=|\mathbf{h}_{u}-\mathbf{h}_{v}|$.

\subsubsection{Link Prediction on Multi-relational Graphs}
Different from graph form $G(E,V)$,
the multi-relational graph $G(E,V,R)$ usually contains different types of edges, where $r(u,v)$ indicates the edge type $r\in R$ existing between $u$ and $v$.
For example,
in recommendation system, users and items are nodes of the bipartite graph, and the edge between nodes could be ``click'' and ``add\_to\_cart''.
In the knowledge graph (KG) scenario,
nodes represent real-world entities 
and edges are relations between entities.

Recently, several GLP models have been developed for the LP task on KGs \citep{schlichtkrull2018modeling,vashishth2019composition}.
Based on the MPNN framework in \eqref{eq:mp1} and \eqref{eq:mp2},
R-GCN \citep{schlichtkrull2018modeling} proposes to model different 
edge types
through edge-specific weight matrix $\mathbf{W}_r^k$ for $r\in R$, where the MPNN in R-GCN is instantiated as:
\begin{equation}
\label{eq:rgcn}
	\mathbf{h}_{v}^{k+1} = ACT_k \big(
	\sum\nolimits_{r(u,v)} \mathbf{W}_r^{k+1}\mathbf{h}_{u}^k.
	\big)
\end{equation}


Similar to R-GCN modeling, D-GCN \citep{marcheggiani2017encoding} and W-GCN \citep{shang2019end} are also restricted to learning embeddings for nodes only.
Instead, 
CompGCN \citep{vashishth2019composition} propose to jointly learn entity and relation embeddings:
\begin{equation}
\label{eq:compgcn}
	\mathbf{h}_{v}^{k+1} = ACT_k \big(
	\sum\nolimits_{r(u,v)} \mathbf{W}_{\lambda(r)}^{k+1}\phi(\mathbf{h}_{u}^k,\mathbf{h}_{r}^k)
	\big)
\end{equation}
where 
$\mathbf{h}_{r}^k$ denotes the embedding vector for the specific edge type $r$, and 
$\lambda(r) \in[incoming,$ $outgoing,self\_loop]$
records information of directed edges.
$\phi:\mathbb{R}^d\times\mathbb{R}^d\rightarrow\mathbb{R}^d$ can be any entity-relation composition operation, such as $sub$ in TransE \citep{bordes2013translating}. 

Compared with earlier approaches, GLP models bring remarkable performance gains, illustrating their superiority over LP task. 
However, exiting GLP models rely on manual and empirical graph neural architecture design, 
such as selecting proper $ACT(\cdot)$ and $\phi(\cdot)$.
Thus, a data-aware GLP model is desired.


\begin{figure}[!htbp]
	\centering
	\subfigure[Intra-layer MPNN search.]
	{\includegraphics[width=0.68\linewidth]{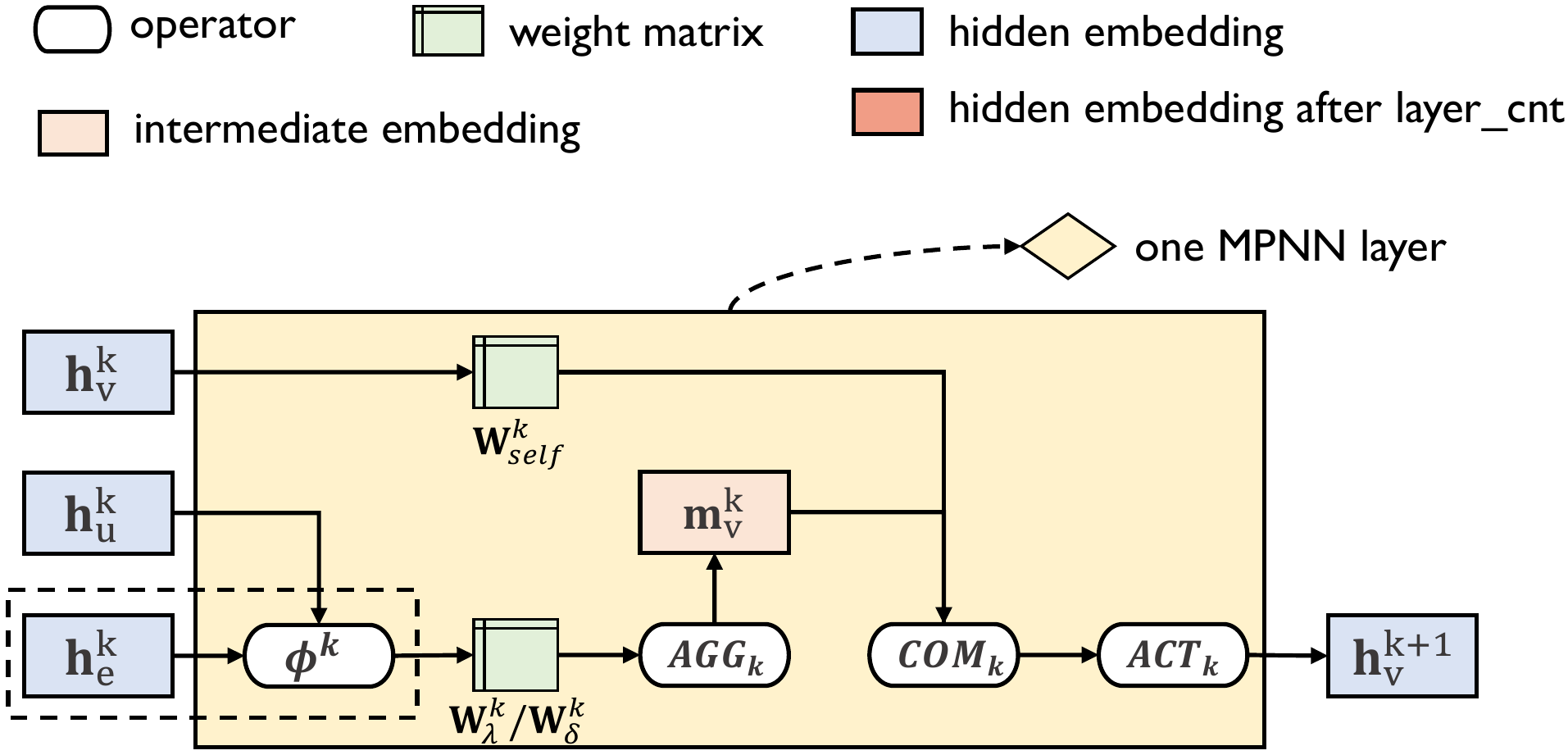}}
	\quad
	\subfigure[Inter-layer MPNN search.]
	{\includegraphics[width=0.28\linewidth]{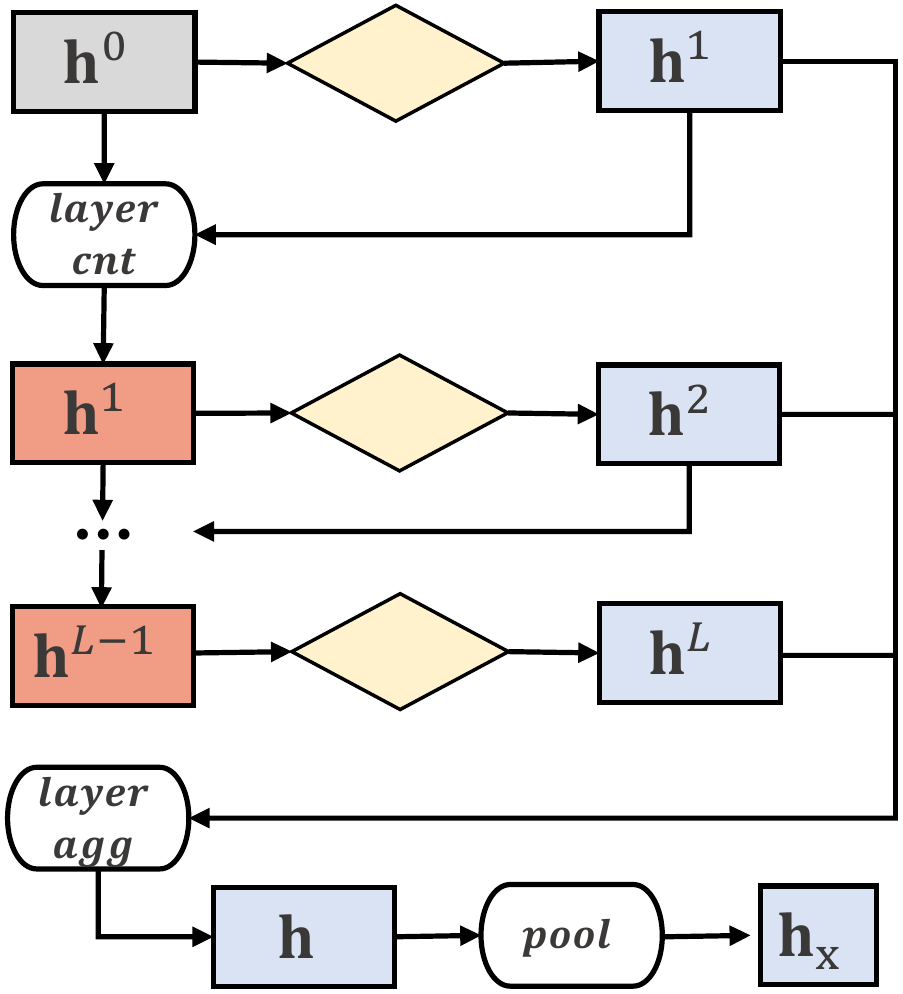}}
	\vspace{-10px}
	\caption{
	The illustration to AutoGEL's search space: a) given representation $\mathbf{h}^k$ in $k$-th layer (including edge embedding $\mathbf{h}_e^k$ if available), AutoGEL searches for proper operators of $\phi^k$, $\mathbf{W}^k$, $AGG_k$, $COM_k$, $ACT_k$ in the intra-layer space, then outputs the hidden node representation $\mathbf{h}^{k+1}_v$. Note that the dotted area will be activated in the scenario of multi-relational graphs.
	}
	\label{fig:search_sapce}
	\vspace{-10px}
\end{figure}

\section{AutoGEL: AutoGNN with Explicit Link Information}

\subsection{Search Space}
\label{sec:3.1}
In this subsection,
AutoGEL 
explicitly models the link information in the MPNN space, and
proposes several novel operators at
both
intra-layer (i.e., the message passing in a specific layer) and inter-layer designs (i.e., the message passing across layers).
The overflow of space is shown in Fig.~\ref{fig:search_sapce}.

\subsubsection{Intra-layer Message Passing Design}
\label{sssec:intra}
As discussed in Sec.~\ref{sec:intro} and Sec.~\ref{ssec:autognn}, 
existing AutoGNNs 
lack modeling of links,
which only utilize the pure link information 
of the node neighborhood.
Such implicit way fails to handle and leverage the complex link information.
To solve this issue,
we present a novel intra-layer message passing framework,
which
instantiates \eqref{eq:mp1} as:
\begin{align}
\label{eq:agg_homo}
& \mathbf{m}_{v}^{k+1} = AGG_k(\{\mathbf{W}_{\delta(u)}^k\mathbf{h}_u^k: u\in N(v)\}),\\
\label{eq:agg_heter}
& \mathbf{m}_{v}^{k+1} = AGG_k(\{\mathbf{W}_{\lambda(e)}^k{\phi^k}(\mathbf{h}_u^k, 
\mathbf{h}_e^k
): u\in N(v)\}), \\
\label{eq:com_shared}
& \mathbf{h}_{v}^{k+1} = ACT_k(COM_k(\{\mathbf{W}_{self}^k\mathbf{h}_{v}^k,\mathbf{m}_{v}^{k+1}\})),
\end{align}
where
\eqref{eq:agg_homo}
and
\eqref{eq:agg_heter}
aggregate neighbor information for homogeneous and multi-relational graphs, respectively.
$\mathbf{W}^k_{\delta(u)}$
and $\mathbf{W}^k_{\lambda(u)}$ encode parts of the link information
as discussed in the following paragraph.
For 
multi-relational
graphs, neighboring nodes from different edge types should impose different influence for the center node during message passing.
Thus,
we 
utilize the edge embedding $\mathbf{h}_e^k$ in \eqref{eq:agg_heter} to further encode the type of links,
where $\mathbf{h}_e^k$ is updated by $\mathbf{h}_{e}^{k+1} = \mathbf{W}_{rel}^{k}\mathbf{h}_{e}^{k}$.
And we incorporate the composition operator $\phi(\cdot)$
to encode the relationship between edge embedding $\mathbf{h}_{e}^k$
with node embedding $\mathbf{h}_{u}^k$.
We highlight the differences between 
the standard MPNN space (see \eqref{eq:mp1} in Sec. 2.2)
with \eqref{eq:agg_homo}, \eqref{eq:agg_heter} and \eqref{eq:com_shared} as follows:


\begin{itemize}[leftmargin=*]
\item 
\textbf{Linear Transformation} $\mathbf{W}^k$: Given the hidden representation from the last layer, we first apply linear transformation towards embeddings.
In 
\eqref{eq:agg_homo},
we assign neighborhood-type specific weight matrices $\mathbf{W}_{\delta(u)}^k$, where $\delta(u) \in \{self, neigh\}$. $\mathbf{W}_{self}^k$, $\mathbf{W}_{neigh}^k$ are introduced for the node itself and neighbors respectively.
This is a weak attention mechanism towards the basic link information for homogeneous graphs, which
can distinguish edges between self-type and
neighbor-type.
In multi-relational graphs, 
we use edge-aware filters $\mathbf{W}^k_{\lambda(e)}$, where $\lambda(e) \in \{self\_loop, original, inverse\}$ encodes the direction information of edge $e$. We use $\mathbf{W}^k_{sl}$, $\mathbf{W}^k_O$, $\mathbf{W}^k_I$ for self-loop, original, and inverse edge separately. 

\item \textbf{Composition Operator} $\phi^k$ and \textbf{Edge Embedding} $\mathbf{h}_{e}^k$: 
Following CompGCN \citep{vashishth2019composition}, we utilize the composition operator $\phi(\mathbf{h}_{u}^k,\mathbf{h}_{e}^k)$ to capture message between the node and edge embeddings before aggregation step. 
While CompGCN empirically selects the most proper
$\phi(\cdot)$ through grid search, we introduce this novel design dimension into our search space, so that 
AutoGEL
is able to search for the most suitable one together with other design dimensions in a more efficient way. 
Specifically, we incorporate the candidate operators $\{sub, mult, corr\}$, where
$sub(\cdot)=\mathbf{h}_{u}^k-\mathbf{h}_{e}^k$ \citep{bordes2013translating}, 
$mult(\cdot)=\mathbf{h}_{u}^k \ast \mathbf{h}_{e}^k$ \citep{yang2014embedding}, 
$corr(\cdot)=\mathbf{h}_{u}^k\star \mathbf{h}_{e}^k$ \citep{nickel2016holographic}.
Together with $\mathbf{W}_{\lambda(e)}^k$,
the search design on
$\phi(\mathbf{h}_u^k,\mathbf{h}_e^k)$ enables AutoGEL to capture semantic meaningful edges by $\mathbf{h}_e^k$, and the interaction between nodes with edges by $\phi(\cdot)$.
That is why AutoGEL can handle the LP task on multi-relational graphs, while another edge-level model~\citep{you2020design} cannot (see Tab.~\ref{tab:autognn}).
 
\end{itemize}

\subsubsection{Inter-Layer Message Passing Design}
\label{sssec:inter}

Traditional MPNNs follows 
the way in \eqref{eq:mp1},
i.e., the input of each MPNN layer is the output of last layer.
Motivated by \citep{xu2018representation, li2019deepgcns},
it is beneficial to use the combination of previous layers as input to each layer.
In this paper,
we also design 
the inter-layer search space to enables the flexible and powerful GNN architecture to be searched.
Specifically, we provide two design dimensions: layer connectivity and layer aggregation.

\begin{itemize}[leftmargin=*]
\item \textbf{Layer Connectivity:} 
The literature \citep{li2019deepgcns} have shown that incorporating skip connections (i.e., residual connections and dense connections) across MPNN layers can help alleviating the over-smoothing issue \citep{li2018deeper} and empirically improve the model performance.
In this work, we conduct systematical investigation towards the joint impact of skip connections together with other design dimension. We select the way of layer connectivity from the set $\{skip, lc\_sum, lc\_concat\}$. 
Moreover, the layer connectivity function is given as:
\begin{equation}
\label{eq:layer_connect}
\mathbf{h}^{k+1}
\leftarrow
layer\_cnt(\mathbf{h}^k,\mathbf{h}^{k+1})=\left\{
\begin{aligned}
&\mathbf{h}^{k+1}, \ \ skip, \\
&sum(\mathbf{h}^k,\mathbf{h}^{k+1}), \ \ lc\_sum, \\
&\mathbf{W}concat(\mathbf{h}^k,\mathbf{h}^{k+1}), \ \  lc\_concat,
\end{aligned}
\right.
\end{equation}
where 
$\mathbf{h}^k$ 
denotes embeddings output from $k$-th MPNN intra-layer.
As shown in Fig.~\ref{fig:search_sapce}~(b),
the representation $\mathbf{h}^k$ will be fed into
$k+1$-th MPNN layer to learn $\mathbf{h}^{k+1}$.
Then, AutoGEL
combines $\mathbf{h}^k$ with $\mathbf{h}^{k+1}$ as in \eqref{eq:layer_connect}
to form a new representation,
which will be fed to the next layer.
Note that 
another AutoGNN
SANE \cite{zhao2021search} does not include $lc\_concat$.


\item \textbf{Layer Aggregation:} JKNet \citep{xu2018representation} shows that the layer-wise aggregation allows the adaptive representation learning. 
The set of candidate layer-wise aggregation 
defined in AutoGEL
is $\{skip, la\_concat, la\_max\}$. 
Specifically, the layer aggregation function is defined as:
\begin{equation}
\label{layer_agg}
\mathbf{h}=layer\_agg(\mathbf{h}^1,\dots,\mathbf{h}^{L})=\left\{
\begin{aligned}
&\mathbf{h}^{L}, \ \ skip, \\
&[\mathbf{h}^1||\dots||\mathbf{h}^{L}], \ \ la\_concat, \\
&max(\mathbf{h}^1,\dots,\mathbf{h}^{L}), \ \ la\_max.
\end{aligned}
\right.
\end{equation}
Note that $layer\_agg$ aggregates the representations generated from MPNN layers, i.e., those that have not been processed by $layer\_cnt$ (Fig.~\ref{fig:search_sapce}~(b)).
And
previous AutoGNNs
NAS-GCN  \citep{jiang2020graph} and  \citep{you2020design}
do not include this operator 
$layer\_agg$.
\end{itemize}

\subsubsection{Pooling}
\label{sssec:pool}

After the intra-layer (Sec.~\ref{sssec:intra}) and inter-layer (Sec.~\ref{sssec:inter})
message passing stages,
pooling operation 
$\mathbf{h}_{x} = R(\{\mathbf{h}_v | v\in \mathcal{X}\})$
induces high-level representations,
where $x$ and $\mathcal{X}$ depend on
the given task.

For LP task on homogeneous graphs,
the pooling operation
outputs
the representations of
links.
In SEAL \citep{zhang2018link}, subgraph-level sortpooling method is utilized to readout information from the entire enclosing subgraph for the target link. 
It is proved in \citep{srinivasan2019equivalence} that joint prediction task only requires joint structure representations of target node-set $S$.
Thus, it is not necessary to introduce
complex subgraph-level pooling methods.
We
following DE-GNN \citep{li2020distance} 
to learn 
link representations by readout only from target node-set, 
i.e., $\mathbf{h}_{e} = R(\{\mathbf{h}_v | v\in S\})$.
Note that the original setting difference-pooling for $R(\cdot)$ in DE-GNN
does not achieve competitive performance in the empirical study.
Instead, we provide the powerful pooling operations $\{sum, max, concat\}$ to be selected for $R(\cdot)$.
As for multi-relational graphs, the pooling stage is not required since 
the edge embedding $\mathbf{h}_{e}$ is learned in the intra-layer MPNN.


For the node classification task,
AutoGEL simply removes $R(\cdot)$ from the search space as literature does.
For the graph classification task, 
the pooling operation outputs high-level graph representation
$\mathbf{h}_{G} = R(\{\mathbf{h}_v | v\in G\})$, and $R(\cdot) \in \{global\_add\_pool, global\_mean\_pool, global\_max\_pool\}$.

%

\subsection{
	Search Algorithm
	}
\label{ssec:snas}

Given a candidate set $\mathcal{O}$ for a specific operator (e.g., 
$\{sum, mean, max\}$ for $AGG_k$), let $\mathbf{x}$ be the hidden vector to be fed into this operator,
and $\alpha_{o}$ records the weight that operation $o \in \mathcal{O}$ to be selected.
Then the output from this operator is computed as
$
\bar{o}(\mathbf{x})
=
\sum_{o\in\mathcal{O}}
\theta_{o}\cdot o(\mathbf{x})
$, 
where $\theta_{o}\in\{0,1\}$.
There are multiple operators in AutoGEL's space (see Sec.~\ref{sec:3.1}),
including intra-layer level operators (i.e., $\phi^k,\mathbf{W}^k,AGG_k,COM_k,ACT_k$),
inter-layer operators
(i.e., $layer\_cnt, layer\_agg$), and pooling operator $R(\cdot)$.
Let $\bm{\theta}$ denote the operation selection for all operators.
The GNN search problem can be formulated as $\max_{\bm{\theta},\bm{\omega}} f(\bm{\theta},\bm{\omega};D)$, where $f(\cdot)$ evaluates the performance of a GNN model $\bm{\theta}$ with weight $\bm{\omega}$ on the graph data $D$.

As discussed in Sec.~\ref{ssec:autognn},
the search algorithm adopted in existing AutoGNNs suffers from several issues.
Especially,
the most similar prior work SANE \citep{zhao2021search}
adopts DARTS \citep{liu2018darts},
which directly relaxes $\bm{\theta}$ to be continuous
and makes the objective $f(\bm{\theta},\bm{\omega};D)$ deterministic differentiable.
However, several drawbacks brought by the mixed strategy of DARTS \citep{liu2018darts} have been observed and discussed in the community of neural architecture search.
The mixed strategy usually leads to the inconsistent performance issue, i.e., the performance of the derived child network at the end of the searching stage shows significant degradation compared with the performance of the parent network before architecture derivation.
That is because
the relaxed $\bm{\theta}$ 
cannot converge to a one-hot vector \citep{zela2019understanding,chu2020fair},
thus removing those operations at the end of search 
actually lead to a different architecture from the final searching result.
Moreover, the mixed strategy must maintain all operators in the whole supernet, which requires more computational resources than the one-hot vector \citep{yao2020efficient}. 

Fortunately,
SNAS \citep{xie2018snas}
leverages the concrete distribution \citep{maddison2016concrete,jang2016categorical}
to propose a stochastic differentiable 
algorithm, 
which enables the search objective differentiable with the reparameterization trick.
Let a GNN model $\bm{\theta}$ being sampled from the distribution $p_{\bm{\alpha}}(\bm{\theta})$ that parameterized by the structure parameter $\bm{\alpha}$ as:
\begin{equation}
\label{eq:reparameterization}
\theta_{o} = 
\frac{\exp((\log \alpha_o - \log(-\log(U_o)))/ \tau)}
{
	\sum_{o'\in \mathcal{O}}
	\exp((\log \alpha_{o'} -\log(-\log(U_{o'})))/ \tau)
},
\end{equation}
where $\tau$ is the temperature of softmax,
and $U_o$ is sampled from the uniform distribution, i.e., $U_o\sim Uniform(0,1)$.
It has been proven that $p(\lim_{\tau \rightarrow 0} \theta_{o} = 1)
= \alpha_{o}/\sum_{o'\in \mathcal{O}}\alpha_{o'}$ \citep{maddison2016concrete}.
This first guarantees that the probability of $o$ being sampled (i.e., $\theta_{o}=1$)
is proportional to its weight $\alpha_o$.
Besides, the one-hot property $\lim_{\tau \rightarrow 0} \theta_{o} = 1$
makes the stochastic differentiable relaxation unbiased once converged \citep{xie2018snas}.
Then the GNN searching problem is 
reformulated into
$
\max_{\bm{\alpha},\bm{\omega}} \mathbb{E}_{\bm{\theta} \sim p_{\bm{\alpha}}(\bm{\theta})}[f(\bm{\theta},\bm{\omega};D)]
$, where $\mathbb{E}[\cdot]$ is the expectation.
We leverage SNAS \citep{xie2018snas}
to optimize the weight of GNN $\bm{\omega}$ and the weight of operator 
$\bm{\alpha}$.

\section{Experiments}
\label{sec:4}

\subsection{Experimental Setting}
\label{sec:4.1}

AutoGEL\footnote{
Code is available at \url{https://github.com/zwangeo/AutoGEL}
}
is implemented on top of code provided in DE-GNN  \citep{li2020distance} and 
CompGCN \citep{vashishth2019composition} using Pytorch framework \citep{paszke2019pytorch}.
All the experiments are performed using one single RTX 2080 Ti GPU.
More details about 
data sets,
hyper-parameter settings and search space designs
are introduced in 
Appendix~\ref{sssec:dataset},
\ref{sssec:hyper} and \ref{appendix: search_space}, respectively.

\textbf{Task and Data sets.}
For LP task on homogeneous graphs,
we follow \citep{zhang2018link, li2020distance} to utilize the datasets: 
NS \citep{newman2006finding}, 
Power \citep{watts1998collective}, 
Router \citep{spring2002measuring}, 
C.ele \citep{watts1998collective}, 
USAir \citep{batagelj2009pajek}, 
Yeast \citep{von2002comparative} and 
PB \citep{ackland2005mapping}.
As for the LP task on multi-relational graphs,
we mainly adopt benchmark
knowledge graphs (KGs),
FB15k-237 \citep{toutanova2015observed}
and WN18RR \citep{dettmers2018convolutional}.

For the node classification task,
we compare models on three popular citation networks \citep{sen2008collective}, i.e., Cora, CiteSeer, and PubMed.
For the graph classification task,
we adopt four standard benchmarks \citep{yanardag2015deep}: 
1) social network datasets: IMDB-BINARY and IMDB-MULTI, 
2) bioinformatics datasets: MUTAG and PROTEINS.

\textbf{Evaluation Metrics.}
For node classification and graph classification,
we adopt average accuracy as measurement.
We report AUC with standard deviation for LP task on homogeneous graphs. 
For LP on KGs, we adopt standard evaluation matrices:
\vspace{-5px}
\begin{itemize}[leftmargin=*]
\vspace{-5px}
\item Mean Reciprocal Ranking (MRR): $(\sum_{i=1}^{|S|}1/rank_i)/|S|$, where $S$ and $rank_i$ denote test triples and ranking results, respectively
\vspace{-5px}
\item Hits@N: $(\sum_{i=1}^{|S|}\mathbbm{1}(rank_i\leq{N}))/|S|$, where $\mathbbm{1}$ denotes indicator function, and $N\in\{1,3,10\}$.
\end{itemize}

\textbf{Baselines.}
For LP on homogeneous graphs, we use the following approaches as baselines:
1) heuristic methods: 
CN \citep{butun2018extension}, RA \citep{zhou2009predicting}, and Katz \citep{katz1953new},
2) latent feature based methods:
Spectral clustering (SPC) \citep{tang2011leveraging}, LINE \citep{tang2015line} and node2vec (N2V) \citep{grover2016node2vec}, 
3)  GLP methods:
VGAE \citep{kipf2016variational}, PGNN \citep{you2019position}, SEAL \citep{zhang2018link}, and DE-GNN \citep{li2020distance}.

For LP on KGs, we compare AutoGEL with several 
KG embedding approaches:
1) the geometric models: TransE \citep{bordes2013translating} and  RotatE \citep{sun2019rotate},
2) bilinear models: 
DistMult \citep{yang2014embedding} and
ComplEx \citep{trouillon2016complex},
3) GLP models: 
R-GCN \citep{kipf2016variational}, 
SACN \citep{shang2019end}, 
VR-GCN \citep{ye2019vectorized} and 
CompGCN \citep{vashishth2019composition},
4) other NN-based models: 
ConvKB \citep{nguyen2017novel},
ConvE \citep{dettmers2018convolutional}, 
ConvR \citep{jiang2019adaptive} and
HyperER \citep{balavzevic2019hypernetwork}.

For the node classification task,
we compare AutoGEL
with the following baselines:
1) manually designed GNNs: 
GCN \citep{kipf2016semi}, 
GraphSAGE \citep{hamilton2017inductive}, 
GAT \citep{velivckovic2017graph} and 
GIN \citep{xu2018powerful},
2) AutoGNNs: 
GraphNAS \citep{gao2020graph}, 
SANE \citep{zhao2021search} and 
\citep{you2020design}. 
AGNN \cite{zhou2019auto} is not included 
due to no available code.

For the graph classification task,
we compare AutoGEL
with the following baselines:
1) manually designed GNNs: PATCHY-SAN \citep{niepert2016learning}, 
DGCNN \citep{zhang2018end},
GCN \citep{kipf2016semi}, 
GraphSAGE \citep{hamilton2017inductive} and
GIN \citep{xu2018powerful},
2) AutoGNNs:
\citep{you2020design}.
Note that NAS-GCN is specifically designed for molecular property prediction, which is not included in the comparison.


\begin{table}[t]
	\caption{Average AUC (with standard deviation) for LP task on homogeneous graphs}
	\label{tab:result_homo}
	\centering
	\setlength\tabcolsep{3pt}
	\scalebox{0.8}
	{
		\begin{tabular}{c|c|c|c|c|c|c|c|c}
			\toprule
			\bf Type & \bf Model & \bf NS & \bf Power & \bf Router & \bf C.ele & \bf USAir &\bf Yeast & \bf PB \\
			\midrule
			
			\multirow{3}{*}{Heuristic} &CN 
			&94.42$\pm$0.95&58.80$\pm$0.88&56.43$\pm$0.52&85.13$\pm$1.61&93.80$\pm$1.22&89.37$\pm$0.61&92.04$\pm$0.35 \\
			&RA 
			&94.45$\pm$0.93&58.79$\pm$0.88&56.43$\pm$0.51&87.49$\pm$1.41&95.77$\pm$0.92&89.45$\pm$0.62&92.46$\pm$0.37 \\
			&Katz 
			&94.85$\pm$1.10&65.39$\pm$1.59&38.62$\pm$1.35&86.34$\pm$1.89&92.88$\pm$1.42&92.24$\pm$0.61&92.92$\pm$0.35 \\ \midrule
			\multirow{3}{*}{Latent} &SPC 
			&89.94$\pm$2.39&91.78$\pm$0.61&68.79$\pm$2.42&51.90$\pm$2.57&74.22$\pm$3.11&93.25$\pm$0.40&83.96$\pm$0.86 \\
			&LINE 
			&80.63$\pm$1.90&55.63$\pm$1.47&67.15$\pm$2.10&69.21$\pm$3.14&81.47$\pm$10.71&87.45$\pm$3.33&76.95$\pm$2.76 \\
			&N2V 
			&91.52$\pm$1.28&76.22$\pm$0.92&65.46$\pm$0.86&84.11$\pm$1.27&91.44$\pm$1.78&93.67$\pm$0.46&85.79$\pm$0.78 \\ \midrule
			\multirow{4}{*}{GLP} &VGAE 
			&94.04$\pm$1.64&71.20$\pm$1.65&61.51$\pm$1.22&81.80$\pm$2.18&89.28$\pm$1.99&93.88$\pm$0.21&90.70$\pm$0.53 \\
			&PGNN 
			&94.88$\pm$0.77&-&-&78.20$\pm$0.33&-&-&89.72$\pm$0.32 \\
			&SEAL 
			&98.85$\pm$0.47&87.61$\pm$1.57&96.38$\pm$1.45&\underline{90.30$\pm$1.35}&96.62$\pm$0.72&97.91$\pm$0.52&94.72$\pm$0.46 \\
			&DE-GNN 
			&\underline{99.09$\pm$0.79}&\underline{96.68$\pm$0.29}&\underline{98.69$\pm$0.17}&89.37$\pm$0.17&\underline{98.04$\pm$0.66}&\underline{98.59$\pm$0.26}&\underline{94.95$\pm$0.37} \\
			\midrule
			AutoGNN &AutoGEL &\textbf{99.89$\pm$0.06}&\textbf{98.00$\pm$0.21}&\textbf{99.08$\pm$0.28}&\textbf{92.90$\pm$1.02}&\textbf{98.49$\pm$0.45}&\textbf{99.24$\pm$0.10}&\textbf{97.27$\pm$0.15} \\
			\bottomrule
		\end{tabular}
	}
\vspace{-10px}
\end{table}

\begin{table}[t]
	\caption{MRR and Hits@N for LP task on knowledge graphs}
	\label{tab:result_kg}
	\centering
	\setlength\tabcolsep{5pt}
	\scalebox{0.9}{
		\begin{tabular}{c|c|cccc|cccc}
			\toprule
			\bf Type&\multirow{2}{*}{\bf Model} &\multicolumn{4}{c|}{\bf FB15k-237}&\multicolumn{4}{c}{\bf WN18RR}\\
			&&MRR&Hits@10&Hits@3&Hits@1&MRR&Hits@10&Hits@3&Hits@1\\
			\midrule
			\multirow{2}{*}{Geometric} &TransE 
			&.294&.465&-&-&.226&.501&-&-\\ 
			&RotatE 
			&.338&.533&.375&.241&\underline{.476}&\bf.571&\underline{.492} &.428\\ \midrule
			\multirow{2}{*}{Bilinear} &DisMult 
			&.241&.419&.263&.155&.430&.490&.440&.390\\
			&ComplEx 
			&.247&.428&.275&.158&.440&.510&.460&.410\\ \midrule
			\multirow{4}{*}{NN-based} &ConvKB 
			&.243&.421&.371&.155&.249&.524&.417& .057\\
			&ConvE 
			&.325&.501&.356&.237&.430&.520&.440&.400\\
			&ConvR 
			&.350&.528&.385&.261&.475&.537&.489&\underline{.443} \\
			&HyperER 
			&.341&.520&.376&.252&.465&.522&.477&{.436}\\ \midrule
			\multirow{4}{*}{GLP}  &R-GCN 
			&.248&.417&-&.151&-&-&-&-\\
			&SACN 
			&.350 & \bf{.540} & \underline{.390} &.260 & .470 & .540 & .480& .430 \\
			&VR-GCN 
			&.248&.432&.272&.159&-&-&-&-\\
			&CompGCN 
			&\underline{.355}&.535&\underline{.390}&\underline{.264}&\bf .479&.546&\bf .494&\underline{.443} \\ 
			\midrule
			AutoGNN &AutoGEL & \bf .357& \underline{.538} & \bf .391&\bf .266&\bf .479&\underline{.549}& \underline{.492}&\bf .444 \\
			\bottomrule
		\end{tabular}
	}
\vspace{-15px}
\end{table}

\subsection{Comparison with GLP models}
\label{sec:4.2}

The model comparison for LP task on homogeneous graphs
and knowledge graphs have been summarized in Tab.~\ref{tab:result_homo}
and ~\ref{tab:result_kg}, respectively.
As shown in Tab.~\ref{tab:result_homo},
heuristic methods 
perform well on several datasets, but they fail to handle data sets Power and Router. 
Latent feature-based methods
improve the performance on these two data sets
but cannot achieve competitive results on other data sets.
GLP models outperform heuristic methods and latent feature-based methods, showing their superiority towards LP task. 
%
Specifically, DE-GNN is our strongest baseline, 
where vanilla GCN is adopted to learn node representations for all datasets, then link representation is induced by pooling node embeddings in \eqref{eq:degnn}.
However, 
DE-GNN 
fails to handle the data-diversity issue and
cannot consistently achieve leading performance on all data sets.
In this paper,
AutoGEL
first
adopts the pooling way in DE-GNN,
then enables a more flexible way 
to select the most suitable pooling function $R(\cdot)$ (see Sec.~\ref{sssec:pool}) instead of the fixed pooling function in DE-GNN.
Searching the
pooling function and other operators
make AutoGEL
handle the data-diversity issue,
and consistently achieve the state-of-the-art performance for LP task on homogeneous graphs. 
Furthermore,
we demonstrate the model performance of LP task on knowledge graphs
in Tab.~\ref{tab:result_kg}.
Note that the improvements on KGs is not as obvious as that on homogeneous graphs.
In practice, GLP models run longer than Geometric and Bilinear models, which leads to the difficulty of tuning hyper-parameters (see more discussions in Appendix~\ref{sec:search_efficiency}).

Moreover, we present several cases of searched architectures in Appendix~\ref{ssec:case}.
And we show several ablation studies to provide some insights into the AutoGEL space design in Appendix~\ref{sec:ablation_study}, including the impacts of the inter-level design, pooling operator, weight transformation matrices, and edge embedding.

%


\begin{table}[t]
	\caption{Average accuracy (\%) for node classification and graph classification}
	\label{tab:nc_gc}
	\setlength\tabcolsep{2pt}
	\centering
	\scalebox{0.96}{
		\begin{tabular}{c|c|ccc|cccc}
			\toprule
			\multirow{2}{*}{\bf Type} &\multirow{2}{*}{\bf Model} & \multicolumn{3}{|c}{\bf Node Classification} & \multicolumn{4}{|c}{\bf Graph Classification}\\
			&&  Cora &  CiteSeer &  Pubmed &  IMDB-B &  IMDB-M &  MUTAG &  PROTEINS\\
			\midrule
			&PATCHYSAN &-&-&-& 71.00 & 45.20 & \underline{92.60} & 75.90 \\
			&DGCNN &-&-&-& 70.00 & 47.80 & 85.80 & 75.50 \\
			Manual&GCN & 88.11 & 76.66 & 88.58 & 74.00 & 51.90 & 85.60 & 76.00 \\
			GNNs&GraphSAGE & 87.41 & 75.99 & 88.34 & 72.30 & 50.90 & 85.10 & 75.90 \\
			&GAT & 87.19 & 75.18 & 85.73 & -& -& -& -\\
			&GIN & 86.00 & 73.40 & 87.99 & \underline{75.10} & \underline{52.30} & 89.40 & \underline{76.20} \\
			\midrule
			\multirow{4}{*}{AutoGNN}&GraphNAS & 88.40 & 77.62 & 88.96 & -& -& -& -\\
			&SANE & \underline{89.26} & \bf 78.59 & \textbf{90.47} & -& -& -& -\\
			&\cite{you2020design} & 88.50 & 74.90 &- &-& 47.80 &-& 73.90 \\
			\cmidrule{2-9}
			&AutoGEL & \textbf{89.89} & \underline{77.66} & \underline{89.68} & \textbf{81.20} & \textbf{56.80} & \textbf{94.74} & \textbf{82.68}\\
			\bottomrule
		\end{tabular}
	}
	\vspace{-10px}
\end{table}

\subsection{Comparison with AutoGNN models}
\label{sec:4.3}

To compare with other AutoGNNs,
we also demonstrate the performance of AutoGEL on node-level and graph-level tasks.
The empirical comparison on the node classification and graph classification task is shown in Tab.~\ref{tab:nc_gc}. 
AutoGEL shows the great generalization ability towards different graph tasks. 
AutoGEL outperforms all the manually designed GNN baselines and also achieves competitive results with existing AutoGNNs designed specifically for these tasks.
The data sets for node classification task 
usually contains
rich node features.
AutoGEL simplifies the attention mechanism in existing GNNs
for node classification from $a_{uv}$ (see Sec.~\ref{ssec:autognn})
to $\mathbf{W}_{\delta(u)}^k$ in \eqref{eq:agg_homo}, which leads to slightly inadequate performance.
We notice that AutoGEL brings more significant performance gains on the graph classification task.
The data sets for graph classification 
have not 
sufficient node features as those data sets for node classification, which requires effective learning from graph structures.
AutoGEL is more suitable for this task by
learning from edges.

We also compare the search efficiency between AutoGNNs in Tab.~\ref{tab:search_efficiency}.
Statistics for other AutoGNNs are taken from the start-of-the-art SANE \citep{zhao2021search}, which
sets search epochs to 200 for all the AutoGNN baselines.
To reduce search cost in GraphNAS,
SANE and AutoGEL leverages the idea of parameter sharing 
\cite{pham2018efficient}
to avoid repeatedly training weights of different sampled GNN architectures.
Moreover, AutoGEL
adopts a more advanced search algorithm compared with SANE (see Sec.~\ref{ssec:snas}), thereby further reduces the search cost.
We show experimental results of a variant of AutoGEL in search algorithm in Appendix~\ref{sec:ablation_study}.
And more details about search efficiency are reported in Appendix~\ref{sec:search_efficiency}.

\begin{table}[!t]
	\caption{The search time (clock time in seconds) of AutoGNNs on the node classification task.}
	\label{tab:search_efficiency}
	\centering
	\begin{tabular}{c|c|c|c}
		\toprule
		Model & Cora & Citeseer & PubMed \\
		\midrule
		GraphNAS \citep{gao2020graph} & 3240 & 3665 & 5917\\
		SANE \citep{zhao2021search} & 14 & 35 & 54\\
		AutoGEL & 12 & 16 & 19\\
		\bottomrule
	\end{tabular}
\vspace{-5px}
\end{table}

\begin{figure}[!t]
	\centering
	\subfigure[LP task on NS]{
		\includegraphics[width=0.22\linewidth]{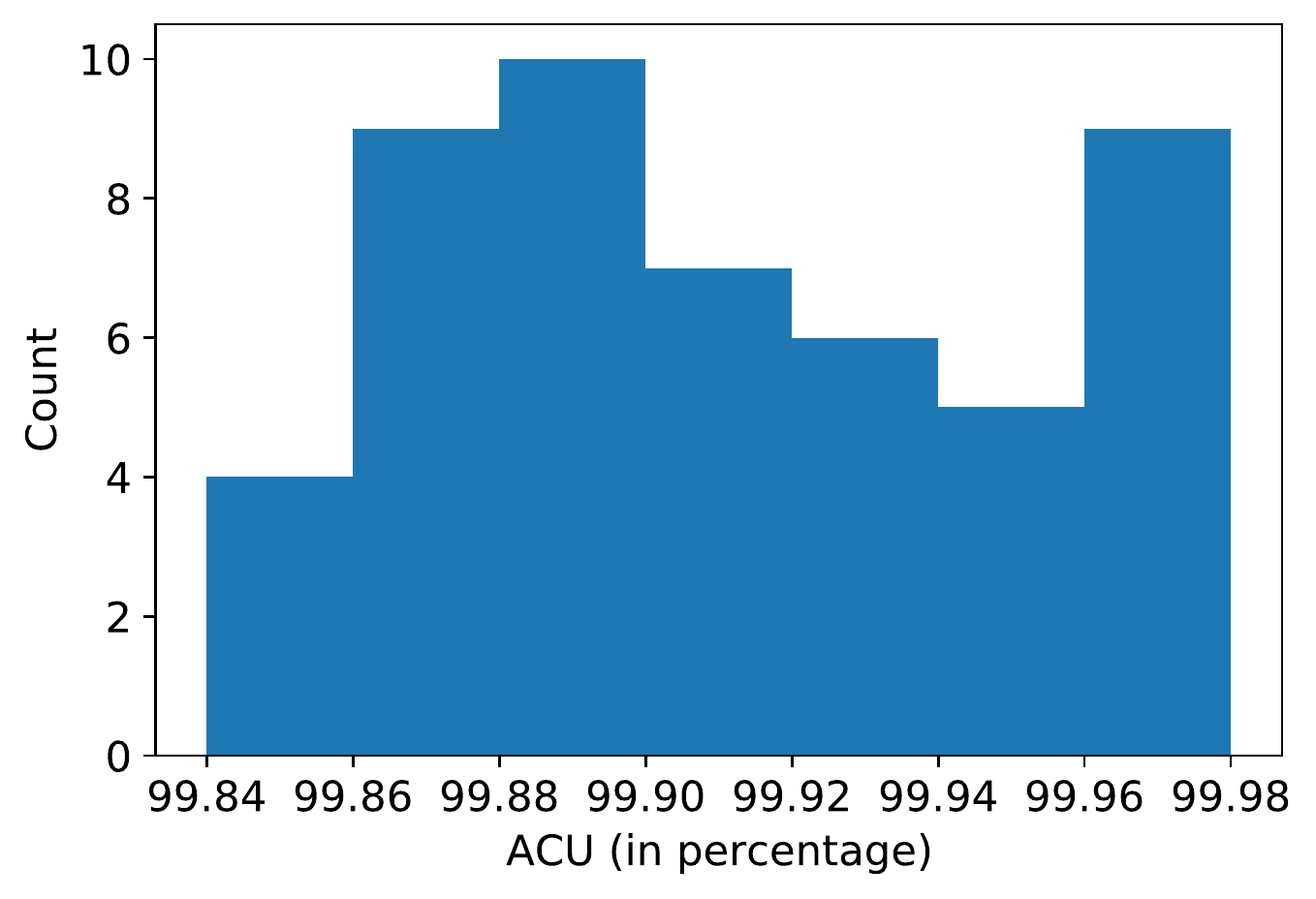}
	}
	\subfigure[LP task on FB15k-237]{
		\includegraphics[width=0.22\linewidth]{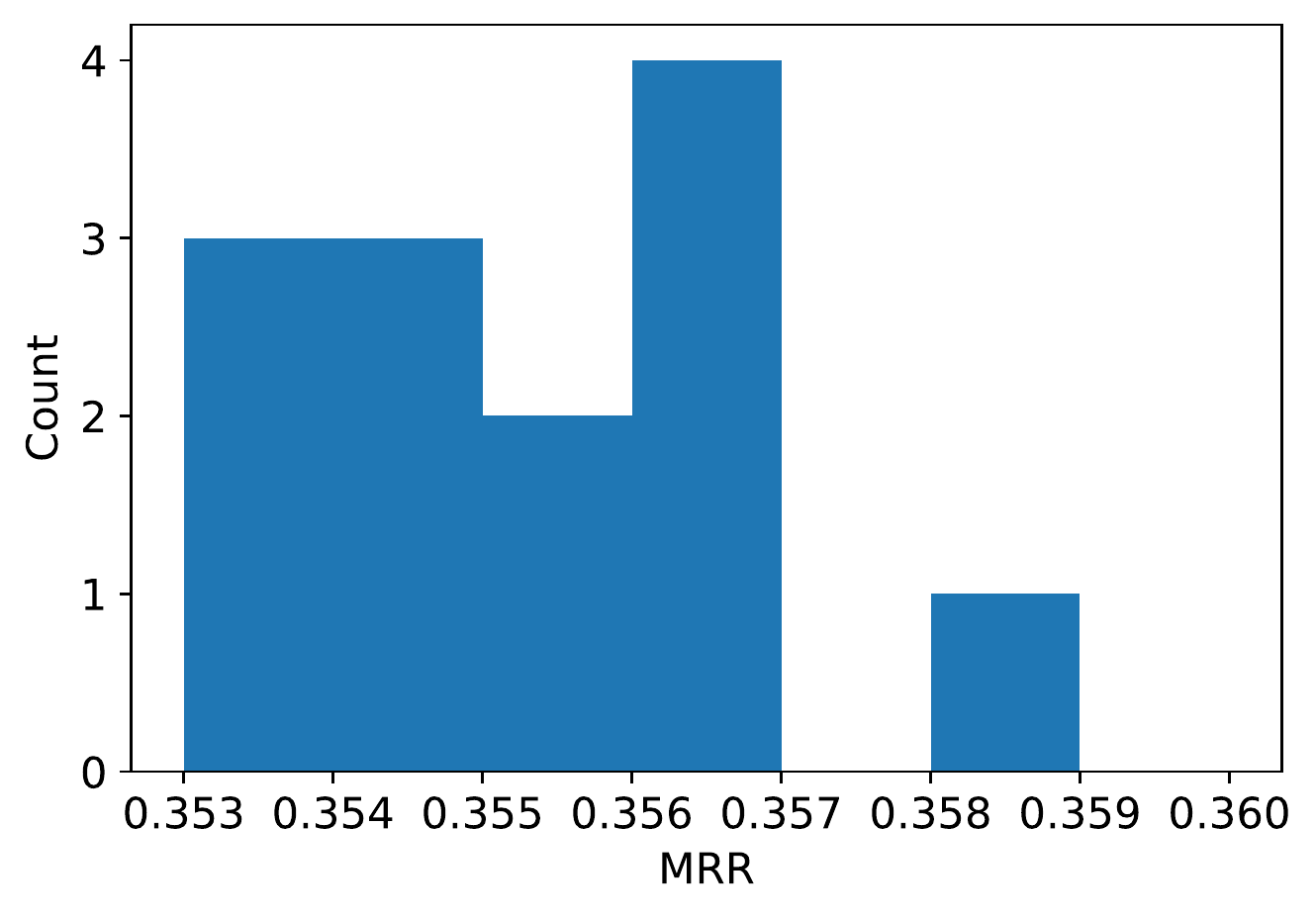}
	}
	\subfigure[NC task on Cora]{
		\includegraphics[width=0.22\linewidth]{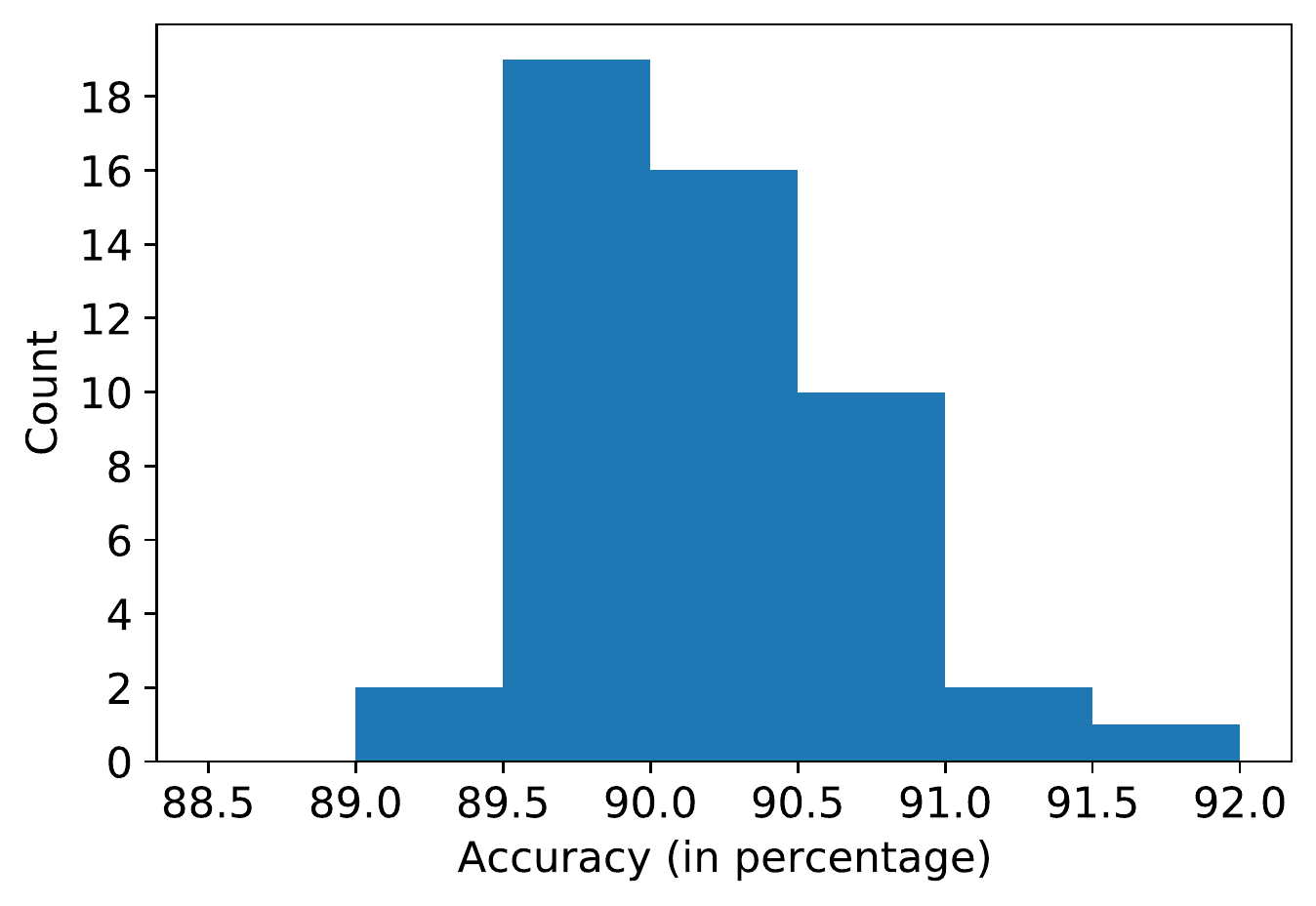}
	}
	\subfigure[GC task on PROT]{
		\includegraphics[width=0.22\linewidth]{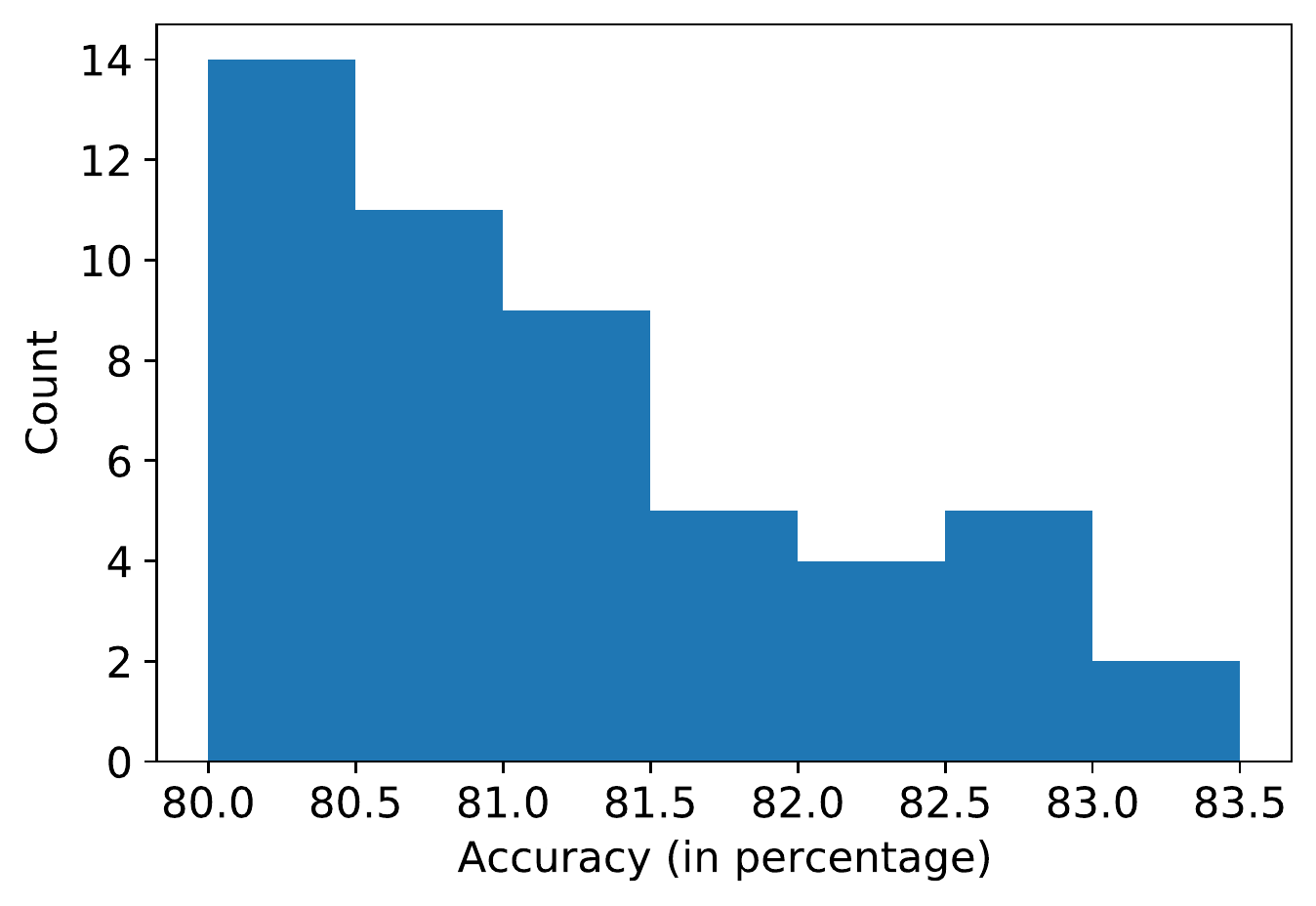}
	}
	\color{teal}
	\caption{Performance frequency statistics over multiple runs for each task}
	\label{fig:frequency_statistics}
\end{figure}

\subsection{The Empirical Study on Robustness}
All effectiveness results in the main context (Tab.~\ref{tab:result_homo}, Tab.~\ref{tab:result_kg}, and Tab.~\ref{tab:nc_gc}) are reported under the average of 4 runs. 
Note that Tab.~\ref{tab:result_kg} and Tab.~\ref{tab:nc_gc} do not contain variance due to space limits. 
To illustrate the robustness of AutoGEL, here we report results after multiple runs of AutoGEL on several tasks in Fig.~\ref{fig:frequency_statistics}.
We set the number of different runs as 10 for FB15k-237 dataset and 50 for the rest, due to the relative longer running time required for FB15k-237. 
We can see that even in the some worst cases, 
AutoGEL still rival or surpass its strongest baselines over all the tasks, its indicating the effectiveness.

\section{Conclusion}
In this paper,
we present a novel AutoGNN with explicit link information, named AutoGEL.
Specifically,
AutoGEL incorporates
the edge embedding in the MPNN space,
and proposes several
novel design dimensions at intra-layer and inter-layer designs.
Moreover,
AutoGEL upgrades the search algorithm of AutoGNNs by leveraging a promising NAS algorithm SNAS.
Experimental results well demonstrate 
that
AutoGEL 
not only achieves the leading performance on the LP task, but also shows competitive results on the node and graph classification tasks.

For future works, one direction worth trying is to adapt AutoGNNs to the LP task on hyper-relational KGs, which contain a lot of hyper-relational facts $r(u_1,\dots,u_n), n\geq 2$.
First, the pioneer data-aware methods for LP tasks on KGs are mainly based on the bilinear models, such as AutoSF~\citep{zhang2020autosf} and ERAS~\citep{shimin2021efficient}.
Introducing the MPNN space can promote a more comprehensive search space because the composition operator is not limited to the bilinear models.
Second, using the multi-relational hypergraph could be a more natural way to model hyper-relational facts \citep{yadati2020neural,di2021searching}.
Another interesting direction is to search GNN architectures on dynamic graph data sets.
Note that search efficiency would be the most challenging issue.
One of the key points is to make full use of previous well-trained GNN controllers.


\section{Acknowledgements}
Lei Chen's work is partially supported by National Key Research and Development Program of China Grant No. 2018AAA0101100, the Hong Kong RGC GRF Project 16202218, CRF Project C6030-18G, C1031-18G, C5026-18G, RIF Project R6020-19, AOE Project AoE/E-603/18, Theme-based project TRS T41-603/20R, China NSFC No. 61729201, Guangdong Basic and Applied Basic Research Foundation 2019B151530001, Hong Kong ITC ITF grants ITS/044/18FX and ITS/470/18FX, Microsoft Research Asia Collaborative Research Grant, HKUST-NAVER/LINE AI Lab, Didi-HKUST joint research lab, HKUST-Webank joint research lab grants.

\bibliographystyle{unsrtnat}
\bibliography{ref.bib}

\begin{thebibliography}{78}
\providecommand{\natexlab}[1]{#1}
\providecommand{\url}[1]{\texttt{#1}}
\expandafter\ifx\csname urlstyle\endcsname\relax
  \providecommand{\doi}[1]{doi: #1}\else
  \providecommand{\doi}{doi: \begingroup \urlstyle{rm}\Url}\fi

\bibitem[Bu et~al.(2018)Bu, Cao, Li, Gao, and Tao]{bu2018gleam}
Zhan Bu, Jie Cao, Hui-Jia Li, Guangliang Gao, and Haicheng Tao.
\newblock Gleam: A graph clustering framework based on potential game
  optimization for large-scale social networks.
\newblock \emph{Knowledge and Information Systems}, 55\penalty0 (3):\penalty0
  741--770, 2018.

\bibitem[Sanchez-Gonzalez et~al.(2018)Sanchez-Gonzalez, Heess, Springenberg,
  Merel, Riedmiller, Hadsell, and Battaglia]{sanchez2018graph}
Alvaro Sanchez-Gonzalez, Nicolas Heess, Jost~Tobias Springenberg, Josh Merel,
  Martin Riedmiller, Raia Hadsell, and Peter Battaglia.
\newblock Graph networks as learnable physics engines for inference and
  control.
\newblock In \emph{ICML}, pages 4470--4479. PMLR, 2018.

\bibitem[Yue et~al.(2020)Yue, Wang, Huang, Parthasarathy, Moosavinasab, Huang,
  Lin, Zhang, Zhang, and Sun]{yue2020graph}
Xiang Yue, Zhen Wang, Jingong Huang, Srinivasan Parthasarathy, Soheil
  Moosavinasab, Yungui Huang, Simon~M Lin, Wen Zhang, Ping Zhang, and Huan Sun.
\newblock Graph embedding on biomedical networks: methods, applications and
  evaluations.
\newblock \emph{Bioinformatics}, 36\penalty0 (4):\penalty0 1241--1251, 2020.

\bibitem[Kipf and Welling(2016{\natexlab{a}})]{kipf2016semi}
Thomas~N Kipf and Max Welling.
\newblock Semi-supervised classification with graph convolutional networks.
\newblock \emph{arXiv preprint arXiv:1609.02907}, 2016{\natexlab{a}}.

\bibitem[Hamilton et~al.(2017)Hamilton, Ying, and
  Leskovec]{hamilton2017inductive}
William~L Hamilton, Rex Ying, and Jure Leskovec.
\newblock Inductive representation learning on large graphs.
\newblock \emph{arXiv preprint arXiv:1706.02216}, 2017.

\bibitem[Vashishth et~al.(2019)Vashishth, Sanyal, Nitin, and
  Talukdar]{vashishth2019composition}
Shikhar Vashishth, Soumya Sanyal, Vikram Nitin, and Partha Talukdar.
\newblock Composition-based multi-relational graph convolutional networks.
\newblock \emph{arXiv preprint arXiv:1911.03082}, 2019.

\bibitem[Li et~al.(2020)Li, Wang, Wang, and Leskovec]{li2020distance}
Pan Li, Yanbang Wang, Hongwei Wang, and Jure Leskovec.
\newblock Distance encoding: Design provably more powerful neural networks for
  graph representation learning.
\newblock \emph{NeurIPS}, 33, 2020.

\bibitem[Niepert et~al.(2016)Niepert, Ahmed, and Kutzkov]{niepert2016learning}
Mathias Niepert, Mohamed Ahmed, and Konstantin Kutzkov.
\newblock Learning convolutional neural networks for graphs.
\newblock In \emph{ICML}, pages 2014--2023. PMLR, 2016.

\bibitem[Zhang et~al.(2018)Zhang, Cui, Neumann, and Chen]{zhang2018end}
Muhan Zhang, Zhicheng Cui, Marion Neumann, and Yixin Chen.
\newblock An end-to-end deep learning architecture for graph classification.
\newblock In \emph{AAAI}, volume~32, 2018.

\bibitem[You et~al.(2020)You, Ying, and Leskovec]{you2020design}
Jiaxuan You, Zhitao Ying, and Jure Leskovec.
\newblock Design space for graph neural networks.
\newblock \emph{NeurIPS}, 33, 2020.

\bibitem[Gilmer et~al.(2017)Gilmer, Schoenholz, Riley, Vinyals, and
  Dahl]{gilmer2017neural}
Justin Gilmer, Samuel~S Schoenholz, Patrick~F Riley, Oriol Vinyals, and
  George~E Dahl.
\newblock Neural message passing for quantum chemistry.
\newblock In \emph{ICML}, pages 1263--1272. PMLR, 2017.

\bibitem[Elsken et~al.(2019)Elsken, Metzen, Hutter, et~al.]{elsken2019neural}
Thomas Elsken, Jan~Hendrik Metzen, Frank Hutter, et~al.
\newblock Neural architecture search: A survey.
\newblock \emph{J. Mach. Learn. Res.}, 20\penalty0 (55):\penalty0 1--21, 2019.

\bibitem[Zhou et~al.(2019)Zhou, Song, Huang, and Hu]{zhou2019auto}
Kaixiong Zhou, Qingquan Song, Xiao Huang, and Xia Hu.
\newblock Auto-gnn: Neural architecture search of graph neural networks.
\newblock \emph{arXiv preprint arXiv:1909.03184}, 2019.

\bibitem[Gao et~al.(2020)Gao, Yang, Zhang, Zhou, and Hu]{gao2020graph}
Yang Gao, Hong Yang, Peng Zhang, Chuan Zhou, and Yue Hu.
\newblock Graph neural architecture search.
\newblock In \emph{IJCAI}, volume~20, pages 1403--1409, 2020.

\bibitem[Jiang and Balaprakash(2020)]{jiang2020graph}
Shengli Jiang and Prasanna Balaprakash.
\newblock Graph neural network architecture search for molecular property
  prediction.
\newblock \emph{arXiv preprint arXiv:2008.12187}, 2020.

\bibitem[Zhao et~al.(2021)Zhao, Yao, and Tu]{zhao2021search}
Huan Zhao, Quanming Yao, and Weiwei Tu.
\newblock Search to aggregate neighborhood for graph neural network.
\newblock \emph{ICDE}, 2021.

\bibitem[Ding et~al.(2021)Ding, Yao, Zhao, and Zhang]{ding2021diffmg}
Yuhui Ding, Quanming Yao, Huan Zhao, and Tong Zhang.
\newblock Diffmg: Differentiable meta graph search for heterogeneous graph
  neural networks.
\newblock In \emph{SIGKDD}, pages 279--288, 2021.

\bibitem[Wei et~al.(2021)Wei, Zhao, Yao, and He]{wei2021pooling}
Lanning Wei, Huan Zhao, Quanming Yao, and Zhiqiang He.
\newblock Pooling architecture search for graph classification.
\newblock \emph{CIKM}, 2021.

\bibitem[Williams(1992)]{williams1992simple}
Ronald~J Williams.
\newblock Simple statistical gradient-following algorithms for connectionist
  reinforcement learning.
\newblock \emph{Machine learning}, 8\penalty0 (3-4):\penalty0 229--256, 1992.

\bibitem[Liu et~al.(2018)Liu, Simonyan, and Yang]{liu2018darts}
Hanxiao Liu, Karen Simonyan, and Yiming Yang.
\newblock Darts: Differentiable architecture search.
\newblock \emph{arXiv preprint arXiv:1806.09055}, 2018.

\bibitem[Xie et~al.(2018)Xie, Zheng, Liu, and Lin]{xie2018snas}
Sirui Xie, Hehui Zheng, Chunxiao Liu, and Liang Lin.
\newblock Snas: stochastic neural architecture search.
\newblock \emph{arXiv preprint arXiv:1812.09926}, 2018.

\bibitem[Schlichtkrull et~al.(2018)Schlichtkrull, Kipf, Bloem, Van Den~Berg,
  Titov, and Welling]{schlichtkrull2018modeling}
Michael Schlichtkrull, Thomas~N Kipf, Peter Bloem, Rianne Van Den~Berg, Ivan
  Titov, and Max Welling.
\newblock Modeling relational data with graph convolutional networks.
\newblock In \emph{European semantic web conference}, pages 593--607. Springer,
  2018.

\bibitem[Zhang et~al.(2019)Zhang, Yao, Shao, and Chen]{zhang2019nscaching}
Yongqi Zhang, Quanming Yao, Yingxia Shao, and Lei Chen.
\newblock Nscaching: simple and efficient negative sampling for knowledge graph
  embedding.
\newblock In \emph{ICDE}, pages 614--625. IEEE, 2019.

\bibitem[Zhang et~al.(2020{\natexlab{a}})Zhang, Yao, and
  Chen]{zhang2020interstellar}
Yongqi Zhang, Quanming Yao, and Lei Chen.
\newblock Interstellar: Searching recurrent architecture for knowledge graph
  embedding.
\newblock \emph{NeurIPS}, 33:\penalty0 10030--10040, 2020{\natexlab{a}}.

\bibitem[Yang et~al.(2016)Yang, Cohen, and Salakhudinov]{yang2016revisiting}
Zhilin Yang, William Cohen, and Ruslan Salakhudinov.
\newblock Revisiting semi-supervised learning with graph embeddings.
\newblock In \emph{ICML}, pages 40--48. PMLR, 2016.

\bibitem[Toutanova and Chen(2015)]{toutanova2015observed}
Kristina Toutanova and Danqi Chen.
\newblock Observed versus latent features for knowledge base and text
  inference.
\newblock In \emph{Proceedings of the 3rd workshop on continuous vector space
  models and their compositionality}, pages 57--66, 2015.

\bibitem[Zhang et~al.(2020{\natexlab{b}})Zhang, Li, Xia, Wang, and
  Jin]{zhang2020revisiting}
Muhan Zhang, Pan Li, Yinglong Xia, Kai Wang, and Long Jin.
\newblock Revisiting graph neural networks for link prediction.
\newblock \emph{arXiv preprint arXiv:2010.16103}, 2020{\natexlab{b}}.

\bibitem[Zhang and Chen(2018)]{zhang2018link}
Muhan Zhang and Yixin Chen.
\newblock Link prediction based on graph neural networks.
\newblock \emph{arXiv preprint arXiv:1802.09691}, 2018.

\bibitem[Perozzi et~al.(2014)Perozzi, Al-Rfou, and Skiena]{perozzi2014deepwalk}
Bryan Perozzi, Rami Al-Rfou, and Steven Skiena.
\newblock Deepwalk: Online learning of social representations.
\newblock In \emph{SIGKDD}, pages 701--710, 2014.

\bibitem[Grover and Leskovec(2016)]{grover2016node2vec}
Aditya Grover and Jure Leskovec.
\newblock node2vec: Scalable feature learning for networks.
\newblock In \emph{SIGKDD}, pages 855--864, 2016.

\bibitem[Kipf and Welling(2016{\natexlab{b}})]{kipf2016variational}
Thomas~N Kipf and Max Welling.
\newblock Variational graph auto-encoders.
\newblock \emph{arXiv preprint arXiv:1611.07308}, 2016{\natexlab{b}}.

\bibitem[Marcheggiani and Titov(2017)]{marcheggiani2017encoding}
Diego Marcheggiani and Ivan Titov.
\newblock Encoding sentences with graph convolutional networks for semantic
  role labeling.
\newblock \emph{arXiv preprint arXiv:1703.04826}, 2017.

\bibitem[Shang et~al.(2019)Shang, Tang, Huang, Bi, He, and Zhou]{shang2019end}
Chao Shang, Yun Tang, Jing Huang, Jinbo Bi, Xiaodong He, and Bowen Zhou.
\newblock End-to-end structure-aware convolutional networks for knowledge base
  completion.
\newblock In \emph{AAAI}, volume~33, pages 3060--3067, 2019.

\bibitem[Bordes et~al.(2013)Bordes, Usunier, Garcia-Duran, Weston, and
  Yakhnenko]{bordes2013translating}
Antoine Bordes, Nicolas Usunier, Alberto Garcia-Duran, Jason Weston, and Oksana
  Yakhnenko.
\newblock Translating embeddings for modeling multi-relational data.
\newblock In \emph{NIPS}, pages 1--9, 2013.

\bibitem[Yang et~al.(2014)Yang, Yih, He, Gao, and Deng]{yang2014embedding}
Bishan Yang, Wen-tau Yih, Xiaodong He, Jianfeng Gao, and Li~Deng.
\newblock Embedding entities and relations for learning and inference in
  knowledge bases.
\newblock \emph{arXiv preprint arXiv:1412.6575}, 2014.

\bibitem[Nickel et~al.(2016)Nickel, Rosasco, and Poggio]{nickel2016holographic}
Maximilian Nickel, Lorenzo Rosasco, and Tomaso Poggio.
\newblock Holographic embeddings of knowledge graphs.
\newblock In \emph{AAAI}, volume~30, 2016.

\bibitem[Xu et~al.(2018{\natexlab{a}})Xu, Li, Tian, Sonobe, Kawarabayashi, and
  Jegelka]{xu2018representation}
Keyulu Xu, Chengtao Li, Yonglong Tian, Tomohiro Sonobe, Ken-ichi Kawarabayashi,
  and Stefanie Jegelka.
\newblock Representation learning on graphs with jumping knowledge networks.
\newblock In \emph{ICML}, pages 5453--5462. PMLR, 2018{\natexlab{a}}.

\bibitem[Li et~al.(2019)Li, Muller, Thabet, and Ghanem]{li2019deepgcns}
Guohao Li, Matthias Muller, Ali Thabet, and Bernard Ghanem.
\newblock Deepgcns: Can gcns go as deep as cnns?
\newblock In \emph{Proceedings of the IEEE/CVF International Conference on
  Computer Vision}, pages 9267--9276, 2019.

\bibitem[Li et~al.(2018)Li, Han, and Wu]{li2018deeper}
Qimai Li, Zhichao Han, and Xiao-Ming Wu.
\newblock Deeper insights into graph convolutional networks for semi-supervised
  learning.
\newblock In \emph{AAAI}, volume~32, 2018.

\bibitem[Srinivasan and Ribeiro(2019)]{srinivasan2019equivalence}
Balasubramaniam Srinivasan and Bruno Ribeiro.
\newblock On the equivalence between positional node embeddings and structural
  graph representations.
\newblock \emph{arXiv preprint arXiv:1910.00452}, 2019.

\bibitem[Zela et~al.(2019)Zela, Elsken, Saikia, Marrakchi, Brox, and
  Hutter]{zela2019understanding}
Arber Zela, Thomas Elsken, Tonmoy Saikia, Yassine Marrakchi, Thomas Brox, and
  Frank Hutter.
\newblock Understanding and robustifying differentiable architecture search.
\newblock \emph{arXiv preprint arXiv:1909.09656}, 2019.

\bibitem[Chu et~al.(2020)Chu, Zhou, Zhang, and Li]{chu2020fair}
Xiangxiang Chu, Tianbao Zhou, Bo~Zhang, and Jixiang Li.
\newblock Fair darts: Eliminating unfair advantages in differentiable
  architecture search.
\newblock In \emph{ECCV}, pages 465--480. Springer, 2020.

\bibitem[Yao et~al.(2020)Yao, Xu, Tu, and Zhu]{yao2020efficient}
Quanming Yao, Ju~Xu, Wei-Wei Tu, and Zhanxing Zhu.
\newblock Efficient neural architecture search via proximal iterations.
\newblock In \emph{AAAI}, volume~34, pages 6664--6671, 2020.

\bibitem[Maddison et~al.(2016)Maddison, Mnih, and Teh]{maddison2016concrete}
Chris~J Maddison, Andriy Mnih, and Yee~Whye Teh.
\newblock The concrete distribution: A continuous relaxation of discrete random
  variables.
\newblock \emph{arXiv preprint arXiv:1611.00712}, 2016.

\bibitem[Jang et~al.(2016)Jang, Gu, and Poole]{jang2016categorical}
Eric Jang, Shixiang Gu, and Ben Poole.
\newblock Categorical reparameterization with gumbel-softmax.
\newblock \emph{arXiv preprint arXiv:1611.01144}, 2016.

\bibitem[Paszke et~al.(2019)Paszke, Gross, Massa, Lerer, Bradbury, Chanan,
  Killeen, Lin, Gimelshein, Antiga, et~al.]{paszke2019pytorch}
Adam Paszke, Sam Gross, Francisco Massa, Adam Lerer, James Bradbury, Gregory
  Chanan, Trevor Killeen, Zeming Lin, Natalia Gimelshein, Luca Antiga, et~al.
\newblock Pytorch: An imperative style, high-performance deep learning library.
\newblock \emph{arXiv preprint arXiv:1912.01703}, 2019.

\bibitem[Newman(2006)]{newman2006finding}
Mark~EJ Newman.
\newblock Finding community structure in networks using the eigenvectors of
  matrices.
\newblock \emph{Physical review E}, 74\penalty0 (3):\penalty0 036104, 2006.

\bibitem[Watts and Strogatz(1998)]{watts1998collective}
Duncan~J Watts and Steven~H Strogatz.
\newblock Collective dynamics of ‘small-world’networks.
\newblock \emph{nature}, 393\penalty0 (6684):\penalty0 440--442, 1998.

\bibitem[Spring et~al.(2002)Spring, Mahajan, and
  Wetherall]{spring2002measuring}
Neil Spring, Ratul Mahajan, and David Wetherall.
\newblock Measuring isp topologies with rocketfuel.
\newblock \emph{SIGCOMM}, 32\penalty0 (4):\penalty0 133--145, 2002.

\bibitem[Batagelj and Mrvar(2009)]{batagelj2009pajek}
Vladimir Batagelj and Andrej Mrvar.
\newblock Pajek datasets (2006), 2009.

\bibitem[Von~Mering et~al.(2002)Von~Mering, Krause, Snel, Cornell, Oliver,
  Fields, and Bork]{von2002comparative}
Christian Von~Mering, Roland Krause, Berend Snel, Michael Cornell, Stephen~G
  Oliver, Stanley Fields, and Peer Bork.
\newblock Comparative assessment of large-scale data sets of protein--protein
  interactions.
\newblock \emph{Nature}, 417\penalty0 (6887):\penalty0 399--403, 2002.

\bibitem[Ackland et~al.(2005)]{ackland2005mapping}
Robert Ackland et~al.
\newblock Mapping the us political blogosphere: Are conservative bloggers more
  prominent?
\newblock In \emph{BlogTalk Downunder 2005 Conference, Sydney}. BlogTalk
  Downunder 2005 Conference, Sydney, 2005.

\bibitem[Dettmers et~al.(2018)Dettmers, Minervini, Stenetorp, and
  Riedel]{dettmers2018convolutional}
Tim Dettmers, Pasquale Minervini, Pontus Stenetorp, and Sebastian Riedel.
\newblock Convolutional 2d knowledge graph embeddings.
\newblock In \emph{AAAI}, volume~32, 2018.

\bibitem[Sen et~al.(2008)Sen, Namata, Bilgic, Getoor, Galligher, and
  Eliassi-Rad]{sen2008collective}
Prithviraj Sen, Galileo Namata, Mustafa Bilgic, Lise Getoor, Brian Galligher,
  and Tina Eliassi-Rad.
\newblock Collective classification in network data.
\newblock \emph{AI magazine}, 29\penalty0 (3):\penalty0 93--93, 2008.

\bibitem[Yanardag and Vishwanathan(2015)]{yanardag2015deep}
Pinar Yanardag and SVN Vishwanathan.
\newblock Deep graph kernels.
\newblock In \emph{SIGKDD}, pages 1365--1374, 2015.

\bibitem[B{\"u}t{\"u}n et~al.(2018)B{\"u}t{\"u}n, Kaya, and
  Alhajj]{butun2018extension}
Ertan B{\"u}t{\"u}n, Mehmet Kaya, and Reda Alhajj.
\newblock Extension of neighbor-based link prediction methods for directed,
  weighted and temporal social networks.
\newblock \emph{Information Sciences}, 463:\penalty0 152--165, 2018.

\bibitem[Zhou et~al.(2009)Zhou, L{\"u}, and Zhang]{zhou2009predicting}
Tao Zhou, Linyuan L{\"u}, and Yi-Cheng Zhang.
\newblock Predicting missing links via local information.
\newblock \emph{The European Physical Journal B}, 71\penalty0 (4):\penalty0
  623--630, 2009.

\bibitem[Katz(1953)]{katz1953new}
Leo Katz.
\newblock A new status index derived from sociometric analysis.
\newblock \emph{Psychometrika}, 18\penalty0 (1):\penalty0 39--43, 1953.

\bibitem[Tang and Liu(2011)]{tang2011leveraging}
Lei Tang and Huan Liu.
\newblock Leveraging social media networks for classification.
\newblock \emph{Data Mining and Knowledge Discovery}, 23\penalty0 (3):\penalty0
  447--478, 2011.

\bibitem[Tang et~al.(2015)Tang, Qu, Wang, Zhang, Yan, and Mei]{tang2015line}
Jian Tang, Meng Qu, Mingzhe Wang, Ming Zhang, Jun Yan, and Qiaozhu Mei.
\newblock Line: Large-scale information network embedding.
\newblock In \emph{WWW}, pages 1067--1077, 2015.

\bibitem[You et~al.(2019)You, Ying, and Leskovec]{you2019position}
Jiaxuan You, Rex Ying, and Jure Leskovec.
\newblock Position-aware graph neural networks.
\newblock In \emph{ICML}, pages 7134--7143. PMLR, 2019.

\bibitem[Sun et~al.(2019)Sun, Deng, Nie, and Tang]{sun2019rotate}
Zhiqing Sun, Zhi-Hong Deng, Jian-Yun Nie, and Jian Tang.
\newblock Rotate: Knowledge graph embedding by relational rotation in complex
  space.
\newblock \emph{arXiv preprint arXiv:1902.10197}, 2019.

\bibitem[Trouillon et~al.(2016)Trouillon, Welbl, Riedel, Gaussier, and
  Bouchard]{trouillon2016complex}
Th{\'e}o Trouillon, Johannes Welbl, Sebastian Riedel, {\'E}ric Gaussier, and
  Guillaume Bouchard.
\newblock Complex embeddings for simple link prediction.
\newblock In \emph{ICML}, pages 2071--2080. PMLR, 2016.

\bibitem[Ye et~al.(2019)Ye, Li, Fang, Zang, and Wang]{ye2019vectorized}
Rui Ye, Xin Li, Yujie Fang, Hongyu Zang, and Mingzhong Wang.
\newblock A vectorized relational graph convolutional network for
  multi-relational network alignment.
\newblock In \emph{IJCAI}, pages 4135--4141, 2019.

\bibitem[Nguyen et~al.(2017)Nguyen, Nguyen, Nguyen, and Phung]{nguyen2017novel}
Dai~Quoc Nguyen, Tu~Dinh Nguyen, Dat~Quoc Nguyen, and Dinh Phung.
\newblock A novel embedding model for knowledge base completion based on
  convolutional neural network.
\newblock \emph{arXiv preprint arXiv:1712.02121}, 2017.

\bibitem[Jiang et~al.(2019)Jiang, Wang, and Wang]{jiang2019adaptive}
Xiaotian Jiang, Quan Wang, and Bin Wang.
\newblock Adaptive convolution for multi-relational learning.
\newblock In \emph{Proceedings of the 2019 Conference of the North American
  Chapter of the Association for Computational Linguistics: Human Language
  Technologies, Volume 1 (Long and Short Papers)}, pages 978--987, 2019.

\bibitem[Bala{\v{z}}evi{\'c} et~al.(2019)Bala{\v{z}}evi{\'c}, Allen, and
  Hospedales]{balavzevic2019hypernetwork}
Ivana Bala{\v{z}}evi{\'c}, Carl Allen, and Timothy~M Hospedales.
\newblock Hypernetwork knowledge graph embeddings.
\newblock In \emph{International Conference on Artificial Neural Networks},
  pages 553--565. Springer, 2019.

\bibitem[Veli{\v{c}}kovi{\'c} et~al.(2017)Veli{\v{c}}kovi{\'c}, Cucurull,
  Casanova, Romero, Lio, and Bengio]{velivckovic2017graph}
Petar Veli{\v{c}}kovi{\'c}, Guillem Cucurull, Arantxa Casanova, Adriana Romero,
  Pietro Lio, and Yoshua Bengio.
\newblock Graph attention networks.
\newblock \emph{arXiv preprint arXiv:1710.10903}, 2017.

\bibitem[Xu et~al.(2018{\natexlab{b}})Xu, Hu, Leskovec, and
  Jegelka]{xu2018powerful}
Keyulu Xu, Weihua Hu, Jure Leskovec, and Stefanie Jegelka.
\newblock How powerful are graph neural networks?
\newblock \emph{arXiv preprint arXiv:1810.00826}, 2018{\natexlab{b}}.

\bibitem[Pham et~al.(2018)Pham, Guan, Zoph, Le, and Dean]{pham2018efficient}
Hieu Pham, Melody Guan, Barret Zoph, Quoc Le, and Jeff Dean.
\newblock Efficient neural architecture search via parameters sharing.
\newblock In \emph{ICML}, pages 4095--4104. PMLR, 2018.

\bibitem[Zhang et~al.(2020{\natexlab{c}})Zhang, Yao, Dai, and
  Chen]{zhang2020autosf}
Yongqi Zhang, Quanming Yao, Wenyuan Dai, and Lei Chen.
\newblock Autosf: Searching scoring functions for knowledge graph embedding.
\newblock In \emph{ICDE}, pages 433--444. IEEE, 2020{\natexlab{c}}.

\bibitem[Shimin et~al.(2021)Shimin, Quanming, ZHANG, and
  Lei]{shimin2021efficient}
DI~Shimin, YAO Quanming, Yongqi ZHANG, and CHEN Lei.
\newblock Efficient relation-aware scoring function search for knowledge graph
  embedding.
\newblock In \emph{ICDE}, pages 1104--1115. IEEE, 2021.

\bibitem[Yadati(2020)]{yadati2020neural}
Naganand Yadati.
\newblock Neural message passing for multi-relational ordered and recursive
  hypergraphs.
\newblock \emph{NeurIPS}, 33, 2020.

\bibitem[Di et~al.(2021)Di, Yao, and Chen]{di2021searching}
Shimin Di, Quanming Yao, and Lei Chen.
\newblock Searching to sparsify tensor decomposition for n-ary relational data.
\newblock In \emph{Webconf}, pages 4043--4054, 2021.

\bibitem[Bergstra et~al.(2013)Bergstra, Yamins, and Cox]{bergstra2013making}
James Bergstra, Daniel Yamins, and David Cox.
\newblock Making a science of model search: Hyperparameter optimization in
  hundreds of dimensions for vision architectures.
\newblock In \emph{ICML}, pages 115--123. PMLR, 2013.

\bibitem[Barth{\'e}l{\'e}my and Amaral(2011)]{barthelemy2011small}
Marc Barth{\'e}l{\'e}my and Luis A~Nunes Amaral.
\newblock Small-world networks: Evidence for a crossover picture.
\newblock In \emph{The Structure and Dynamics of Networks}, pages 304--307.
  Princeton University Press, 2011.

\bibitem[Matlock et~al.(2019)Matlock, Datta, Le~Dang, Jiang, and
  Swamidass]{matlock2019deep}
Matthew~K Matlock, Arghya Datta, Na~Le~Dang, Kevin Jiang, and S~Joshua
  Swamidass.
\newblock Deep learning long-range information in undirected graphs with wave
  networks.
\newblock In \emph{IJCNN}, pages 1--8. IEEE, 2019.

\bibitem[Dehmamy et~al.(2019)Dehmamy, Barab{\'a}si, and
  Yu]{dehmamy2019understanding}
Nima Dehmamy, Albert-L{\'a}szl{\'o} Barab{\'a}si, and Rose Yu.
\newblock Understanding the representation power of graph neural networks in
  learning graph topology.
\newblock \emph{arXiv preprint arXiv:1907.05008}, 2019.

\end{thebibliography}


\section*{Checklist}

\begin{enumerate}[leftmargin=*]
	
	\item For all authors...
	\begin{enumerate}
		\item Do the main claims made in the abstract and introduction accurately reflect the paper's contributions and scope?
		\answerYes{}
		\item Did you describe the limitations of your work?
		\answerYes{} Please see Section~\ref{sec:4.2}.
		\item Did you discuss any potential negative societal impacts of your work?
		\answerNo{We do not think this work would have any negative societal impacts.}
		\item Have you read the ethics review guidelines and ensured that your paper conforms to them?
		\answerYes{}
	\end{enumerate}
	
	\item If you are including theoretical results...
	\begin{enumerate}
		\item Did you state the full set of assumptions of all theoretical results?
		\answerNA{We do not include theoretical analysis. We mainly focus on empirical study in this work.}
		\item Did you include complete proofs of all theoretical results?
		\answerNA{}
	\end{enumerate}
	
	\item If you ran experiments...
	\begin{enumerate}
		\item Did you include the code, data, and instructions needed to reproduce the main experimental results (either in the supplemental material or as a URL)?
		\answerYes{}
		\item Did you specify all the training details (e.g., data splits, hyperparameters, how they were chosen)?
		\answerYes{Please see Appendix~\ref{sssec:dataset} for data splits and Appendix~\ref{sssec:hyper} for hyperparameters.}
		\item Did you report error bars (e.g., with respect to the random seed after running experiments multiple times)?
		\answerYes{Please see Section~\ref{sec:4.2} for standard deviation values.}
		\item Did you include the total amount of compute and the type of resources used (e.g., type of GPUs, internal cluster, or cloud provider)?
		\answerYes{Please see Section~\ref{sec:4.1} for the GPU type. See Section~\ref{sec:4.3} and Appendix~\ref{sec:computation_cost} for the computation cost.}
	\end{enumerate}
	
	\item If you are using existing assets (e.g., code, data, models) or curating/releasing new assets...
	\begin{enumerate}
		\item If your work uses existing assets, did you cite the creators?
		\answerYes{}
		\item Did you mention the license of the assets?
		\answerYes{All the assets used in this work is public.}
		\item Did you include any new assets either in the supplemental material or as a URL?
		\answerNA{}
		\item Did you discuss whether and how consent was obtained from people whose data you're using/curating?
		\answerYes{All the datasets used in this work are public.}
		\item Did you discuss whether the data you are using/curating contains personally identifiable information or offensive content?
		\answerNA{We do not involve such datasets in this work.}
	\end{enumerate}
	
	\item If you used crowdsourcing or conducted research with human subjects...
	\begin{enumerate}
		\item Did you include the full text of instructions given to participants and screenshots, if applicable?
		\answerNA{We do not conduct such research in this work.}
		\item Did you describe any potential participant risks, with links to Institutional Review Board (IRB) approvals, if applicable?
		\answerNA{}
		\item Did you include the estimated hourly wage paid to participants and the total amount spent on participant compensation?
		\answerNA{}
	\end{enumerate}
	
\end{enumerate}

\newpage

\appendix

\begin{table}[!t]
	\caption{Dataset statistics for link prediction task}
	\label{dataset_stat:lp}
	\centering
	\scalebox{0.905}{
		\begin{tabular}{c|c|c|c|c|c|c|c|c|c}
			\toprule
			\multirow{2}{*}{} & \multicolumn{7}{|c}{Homogeneous Graphs} & \multicolumn{2}{|c}{KGs}\\
			\cmidrule{2-8} \cmidrule{9-10} 
			&NS &Power &Router &C.ele &USAir &Yeast &PB & FB15k\-237 & WN18RR \\
			\midrule
			\# Nodes & 1589 & 4941 & 5022 & 297 & 332 & 2375 & 1222 & 14541 & 40943 \\
			\# Edges & 2742 & 6594 & 6258 & 2148 & 2126 & 11693 & 16714 & 310116 & 93003 \\
			\# Edge types & 1 & 1 & 1 & 1 & 1 & 1 & 1 & 237 & 11\\
			\# Relations &-&-&-&-&-&-&-& 237 & 11\\
			Avg. \# Degrees & 3.45 & 2.67 & 2.49 & 14.46 & 12.81 & 9.85 & 27.36 & 21.33 & 2.27\\
			\midrule
			\# Training & 4387 & 10550 & 10012 & 3436 & 3401 & 18708 & 26742 & 272,115 & 86,835\\
			\# Validation & 548 & 1319 & 1251 & 429 & 425 & 2338 & 3342 & 17,535 & 3,034\\
			\# Testing & 548 & 1319 & 1251 & 429 & 425 & 2338 & 3342 & 20,466 & 3,134\\
			\bottomrule
		\end{tabular}
	}
\end{table}

\begin{table}[t]
	\caption{Dataset statistics for node classification and graph classification task}
	\label{dataset_stat:node and graph}
	\centering
	\scalebox{0.9}{
		\begin{tabular}{c|c|c|c|c|c|c|c}
			\toprule
			\multirow{2}{*}{}&\multicolumn{3}{c}{Node Classification} & \multicolumn{4}{|c}{Graph Classification} \\
			\cmidrule{2-4} \cmidrule{5-8} 
			& Cora & CiteSeer & PubMed & IMDB-B &IMDB-M &MUTAG &PROTEINS\\
			\midrule
			\# Graphs & 1 & 1 & 1 & 1000 & 1500 & 188 & 1113 \\
			\# Nodes & 2708 & 3327 & 19717 & 19.8 (Avg.) & 13.0 (Avg.) & 17.9 (Avg.) & 39.1 (Avg.)\\
			\# Edges & 2742 & 6594 & 6258 & 96.53 (Avg.) & 65.94 (Avg.) & 19.79 (Avg.) & 72.82 (Avg.)\\
			\# Edge types & 1 & 1 & 1 & 1 & 1 & 4 & 1 \\
			\# Node Attr. & 1433 & 3703 & 500 & - & - & 3 & 7 \\
			\# Classes & 7 & 6 & 3 & 2 & 3 & 2 & 2 \\
			\bottomrule
		\end{tabular}
	}
\end{table}

\section{Experiments}

\subsection{More Experimental Settings}
\subsubsection{Dataset Details}
\label{sssec:dataset}

We summarize the dataset statistics in Tab.~\ref{dataset_stat:lp} and Tab.~\ref{dataset_stat:node and graph}. 
In terms of dataset splits, for LP task on homogeneous graphs, we follow \citep{li2020distance} to split 80\%, 10\%, 10\% of existing links for training, validation and testing respectively. The same number of negative links are also included through random sampling. During training phase, positive test links are removed to avoid label leakage. 
For KGs, we follow their standard split
as shown in Tab.~\ref{dataset_stat:lp}.
Moreover, we use 60\%, 20\%, 20\% dataset split for node classification as in \citep{zhao2021search}, and 80\%, 10\%, 10\% for graph classification task to keep the same percentage of test split as in \citep{xu2018powerful} for fair comparison.

\subsubsection{Hyperparameter Settings}
\label{sssec:hyper}
We provide detailed hyperparameter settings in Tab.~\ref{tab:hyper_params} for our implementation. 
Hyperparameters are tuned through hyperopt \footnote{https://github.com/hyperopt/hyperopt} \cite{bergstra2013making}. 

\begin{table}[!h]
	\centering
	\caption{List of value / range of hyperparameters in AutoGEL's implementation}
	\label{tab:hyper_params}
	\scalebox{0.92}{
		\begin{tabular}{c|c|c|c|c}
			\toprule
			\multirow{2}{*}{Hyperparameters} & \multicolumn{2}{c|}{Link Prediction} & \multirow{2}{*}{Node Classification} &  \multirow{2}{*}{Graph Classification}\\ 
			\cmidrule{2-3} 
			& Homo. Graphs & KGs & & \\
			\midrule
			Optimizer & Adam & Adam & Adam & Adam\\
			Learning rate & 1e-4 & \{1e-3,1e-4\} & \{1e-3, 5e-3, 1e-4\} & \{1e-2, 1e-3, 1e-4\} \\
			MPNN layers & 2 & \{1, 2\} & 2 & 4 \\
			Batch size & \{64, 128\} & \{128, 256\} & \{64, 256\} & \{32, 128\} \\
			Hidden dimension & 100 & 200 & \{64, 256\} & \{16, 32, 64\}\\
			Dropout & \{0, 0.2\} & \{0, 0.1, 0.2, 0.3\} & \{0, 0.5\} & \{0, 0.5\}
			\\
			Search epoch & 300 & \{200,300\} & \{30, 200\} & \{30, 200\}\\
			\bottomrule
		\end{tabular}
	}
\end{table}


\subsubsection{Search Space}
\label{appendix: search_space}
Apart from our main designs presented in Section \ref{sec:3.1}, AutoGEL also includes several other intra-level design dimensions in the search space:
\begin{itemize}[leftmargin=*]
	\item \textbf{Aggregation} $AGG_k$: We follow the common design in AutoGNN works (please refer to Section \ref{ssec:autognn} for more details) to include $\{sum, mean, max\}$ for neighborhood aggregation.
	\item \textbf{Combination} $COM_k$: We select combination function from $\{sum, concat\}$. We omit $mlp$ combination since we empirically find simpler combination operator $sum$ and $concat$ adopted in our search space already achieves good performance.
	\item \textbf{Activation} $ACT_k$: Empirical observations from \citep{you2020design} shows the superiority of PReLU as the activation function for GNNs. In this work, we restrict our candidate activation functions in $\{ReLU, PReLU\}$. For the LP task on KGs, we follow the alternative setting to use $tanh$ since we empirically found $ReLU$ and $PReLU$ not suitable. 
	
	\item \textbf{Node Labeling}: 
	The node labeling method (e.g., double-radius node labeling (DRNL) \citep{zhang2018link} and distance encoding (DE) \citep{li2020distance}) is an important component towards the success of structural prediction tasks (e.g., link prediction). 
	AutoGEL presets the DE as the node labeling approach for the LP task due to its generality and empirically good performance.
	DRNL can be regarded as a special case for DE, where the differences between them are marginal. 
	Both DE and DRNL work well in practice \citep{li2020distance}. 
	Moreover, we tried to incorporate this design dimension into the search space and enable it to be jointly searched with other architecture components. 
	Unfortunately, sacrificing some search efficiency may not be able to improve the effectiveness because DE is already a powerful technique.
	Out of this concern, AutoGEL presets DE as the node labeling method to better balance between effectiveness and efficiency.

\end{itemize}


\begin{figure}[t]
	\centering
	\includegraphics[width=10cm]{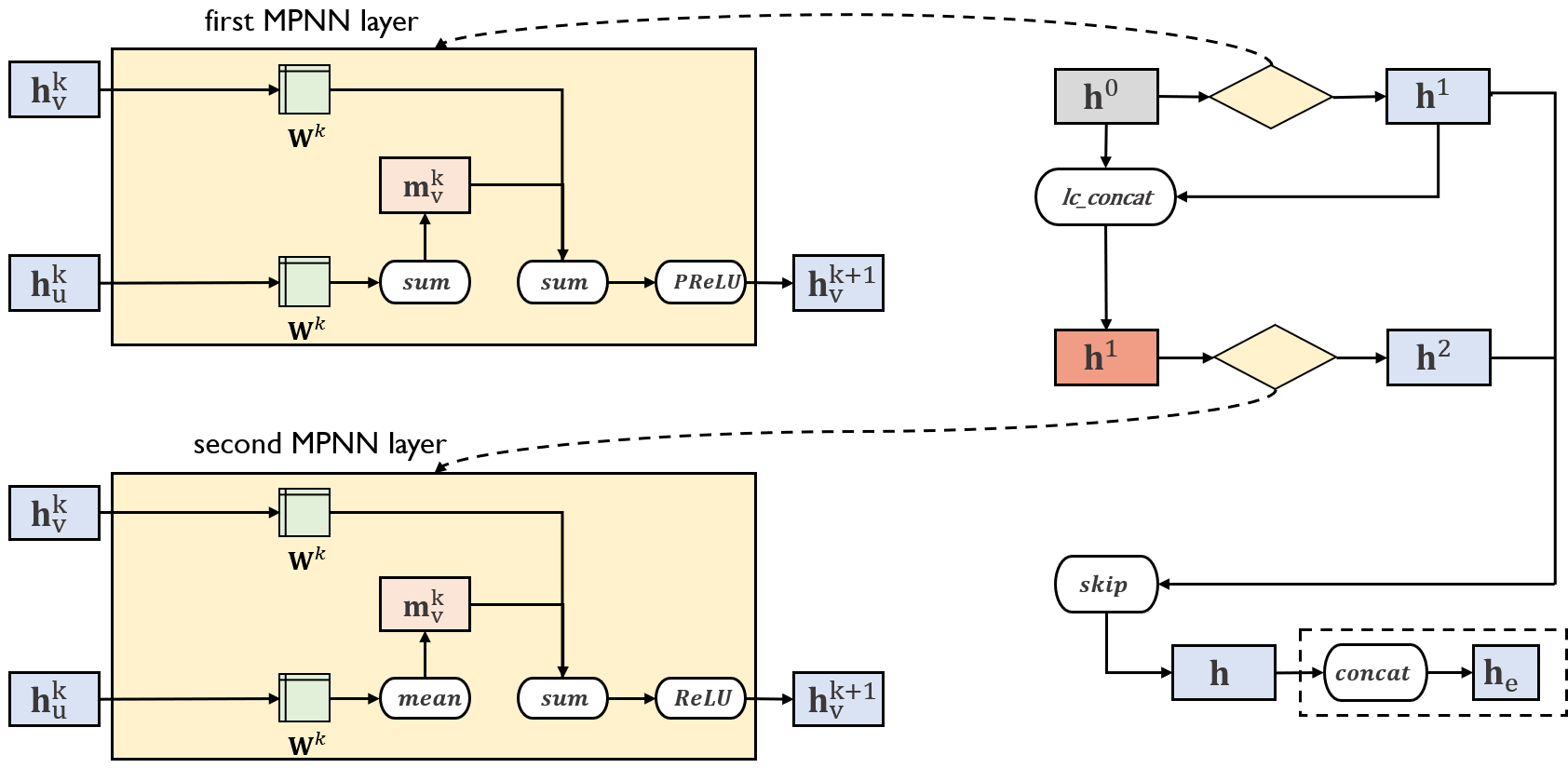}
	\caption{An example: GNN architecture searched by AutoGEL for LP task on PB dataset.}
	\label{fig:searched_pb}
\end{figure}


\begin{figure}[t]
	\centering
	\includegraphics[width=10cm]{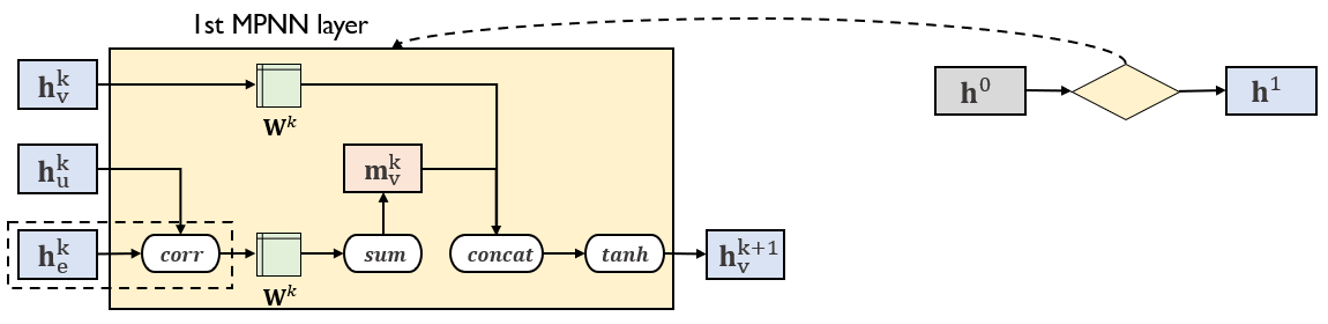}
	\caption{An example: GNN architecture searched by AutoGEL for LP task on FB15k-237 dataset.}
	\label{fig:searched_fb}
\end{figure}

%
%

\subsection{Case Study}
\label{ssec:case}

Here we show some searched architectures for several tasks: link prediction (LP), node classification (NC), and graph classification (GC).

For the LP task (see Fig. \ref{fig:searched_pb} and Fig. \ref{fig:searched_fb}),
we find that the depth of MPNN layers $L$ leading to highest performance is different from graph scenarios.
Generally,
$L=2$ for homogeneous graphs while $L=1$ for knowledge graph.
One possible reason is that KGs are usually more densely connected (see Tab. \ref{dataset_stat:lp} for more dataset details), and deeper MPNN layers would cause the over-smoothing issue, resulting in performance degradation. 
Moreover, it is also discussed in \citep{zhang2018link} that for subgraph-based LP approaches on homogeneous graphs adopted by AutoGEL, 2-hop enclosing subgraphs already contain rich information required for the prediction, therefore $L\textgreater2$ should not be very necessary.

Specifically, for the LP task on KG scenario, we empirically observe that the composition operator $\phi(\mathbf{h}_u,\mathbf{h}_e)$ (see Sec.~\ref{sec:3.1}) should be one of the most critical components.
This operator determines the way how to compose the neighborhood embedding $\mathbf{h}_u$ and edge embedding $\mathbf{h}_e$ to generate the message for the center node $v$.
Actually, the composition operator $\phi$ incorporates the scoring function design in past KG embedding models, such as subtraction for geometric models and multiplication for bilinear models.
From experiments, we observed that $\phi$ is data-dependent. $corr$ is more preferred for the FB15k-237 dataset, and simpler $mult$ is prone to get better results for the WN18RR dataset. 
Using others $\phi$ for these data sets would lead to significantly different performance based on the empirical study.

\begin{figure}[t]
	\centering
	\includegraphics[width=10cm]{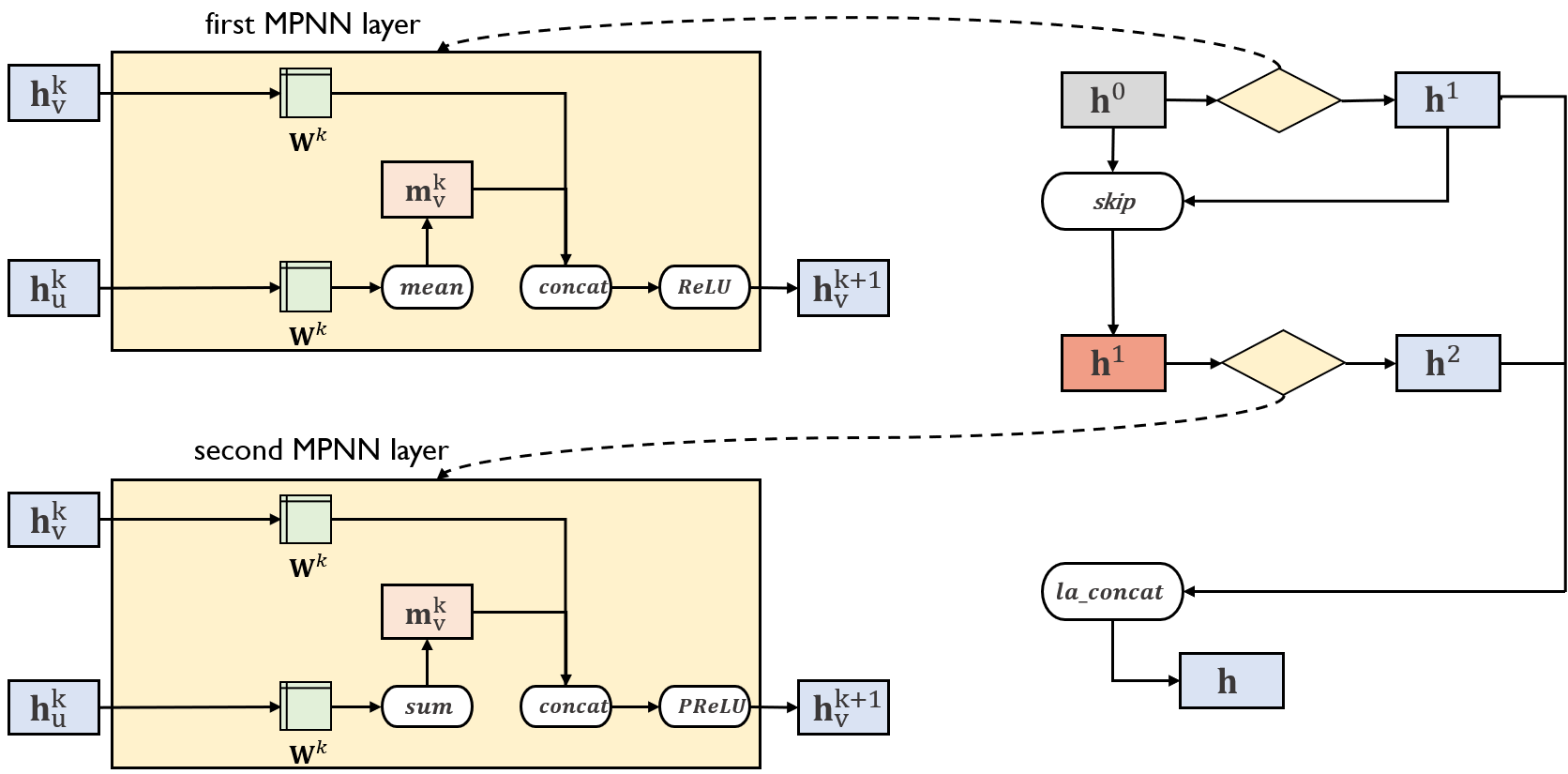}
	\caption{An example: GNN architecture searched by AutoGEL for
	the NC task on PubMed.}
	\label{fig:searched_pubmed}
\end{figure}

For the NC task (see Fig. \ref{fig:searched_pubmed}), pooling operator $R(\cdot)$ is removed from the search space, and we set $L=2$ for all three citation datasets, since we observe performance degradation with larger $L$ on those datasets. 

\begin{figure}[t]
	\centering
	\includegraphics[width=13.5cm]{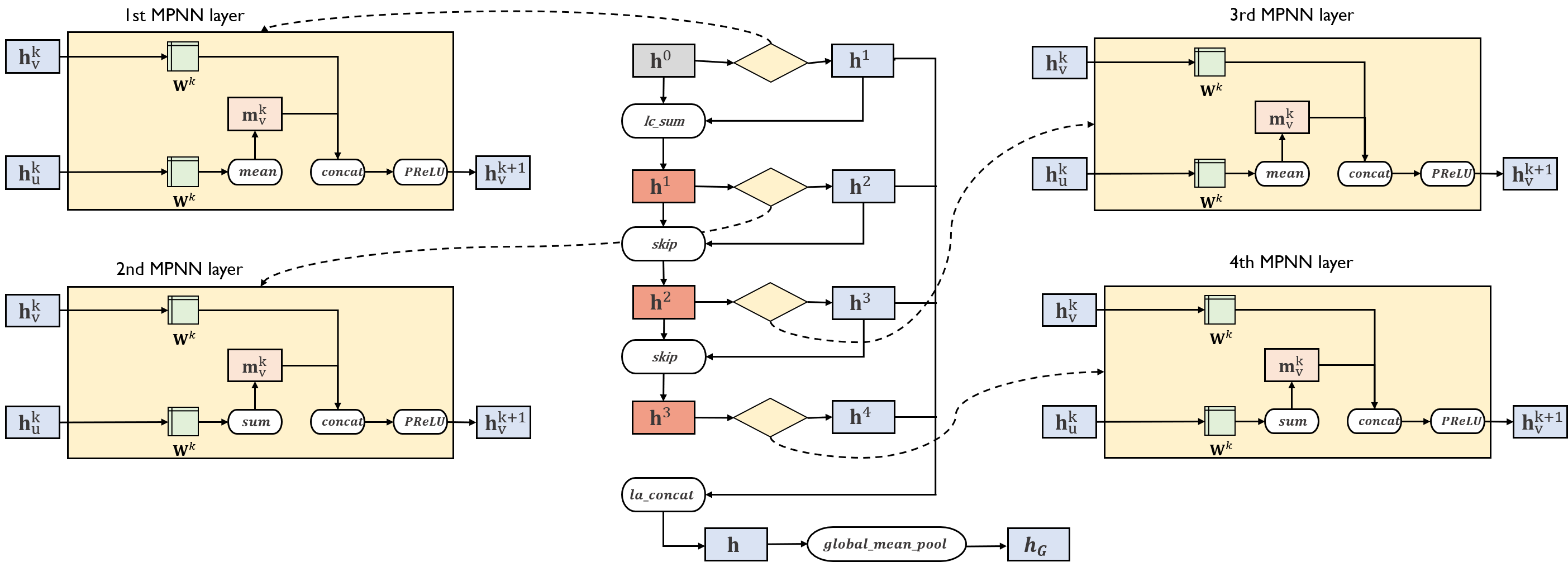}
	\caption{An example: GNN architecture searched by AutoGEL for
	the GC task on PROTEINS}
	\label{fig:searched_PROTEINS}
\end{figure}

For the GC task (see Fig.\ref{fig:searched_PROTEINS}), 
we empirically observe that AutoGEL prefers deeper GNN architectures compared to the LP and NC tasks.
One potential reason is that, citation datasets adopted for the NC task are similar to ``small world'' networks \citep{barthelemy2011small} where each node can reach the entire graph within just a few hops.
But the data sets for the GC task represent graph structures, such as molecules, where deeper architectures might be beneficial to increase effective receptive field.  
Besides, while the NC task mainly relies on local neighborhood (short-range) information, the GC task may require long-range information to capture certain graph properties that are essential to the prediction, such as chemical properties of molecules \citep{matlock2019deep}, and graph moments \citep{dehmamy2019understanding}. Thus deeper GNN architectures are more desired. 

\begin{table}[t]
	\caption{Average AUC (with standard deviation) for LP task on homogeneous graphs}
	\label{tab:ablation_homo}
	\centering
	\setlength\tabcolsep{3pt}
	\scalebox{0.8}
	{
		\begin{tabular}{c|c|c|c|c|c|c|c|c}
			\toprule
			\bf Type & \bf Model & \bf NS & \bf Power & \bf Router & \bf C.ele & \bf USAir &\bf Yeast & \bf PB \\
			\midrule
			
			\multirow{3}{*}{Heuristic} &CN 
			&94.42$\pm$0.95&58.80$\pm$0.88&56.43$\pm$0.52&85.13$\pm$1.61&93.80$\pm$1.22&89.37$\pm$0.61&92.04$\pm$0.35 \\
			&RA 
			&94.45$\pm$0.93&58.79$\pm$0.88&56.43$\pm$0.51&87.49$\pm$1.41&95.77$\pm$0.92&89.45$\pm$0.62&92.46$\pm$0.37 \\
			&Katz 
			&94.85$\pm$1.10&65.39$\pm$1.59&38.62$\pm$1.35&86.34$\pm$1.89&92.88$\pm$1.42&92.24$\pm$0.61&92.92$\pm$0.35 \\ \midrule
			\multirow{3}{*}{Latent} &SPC 
			&89.94$\pm$2.39&91.78$\pm$0.61&68.79$\pm$2.42&51.90$\pm$2.57&74.22$\pm$3.11&93.25$\pm$0.40&83.96$\pm$0.86 \\
			&LINE 
			&80.63$\pm$1.90&55.63$\pm$1.47&67.15$\pm$2.10&69.21$\pm$3.14&81.47$\pm$10.71&87.45$\pm$3.33&76.95$\pm$2.76 \\
			&N2V 
			&91.52$\pm$1.28&76.22$\pm$0.92&65.46$\pm$0.86&84.11$\pm$1.27&91.44$\pm$1.78&93.67$\pm$0.46&85.79$\pm$0.78 \\ \midrule
			\multirow{4}{*}{GLP} &VGAE 
			&94.04$\pm$1.64&71.20$\pm$1.65&61.51$\pm$1.22&81.80$\pm$2.18&89.28$\pm$1.99&93.88$\pm$0.21&90.70$\pm$0.53 \\
			&PGNN 
			&94.88$\pm$0.77&-&-&78.20$\pm$0.33&-&-&89.72$\pm$0.32 \\
			&SEAL 
			&98.85$\pm$0.47&87.61$\pm$1.57&96.38$\pm$1.45&90.30$\pm$1.35&96.62$\pm$0.72&97.91$\pm$0.52&94.72$\pm$0.46 \\
			&DE-GNN 
			&99.09$\pm$0.79&96.68$\pm$0.29&98.69$\pm$0.17&89.37$\pm$0.17&98.04$\pm$0.66&98.59$\pm$0.26&94.95$\pm$0.37 \\
			\midrule
			\multirow{5}{*}{AutoGNN} &AutoGEL &\textbf{99.89$\pm$0.06}&\textbf{98.00$\pm$0.21}&\textbf{99.08$\pm$0.28}&\textbf{92.90$\pm$1.02}&\textbf{98.49$\pm$0.45}&\textbf{99.24$\pm$0.10}&\textbf{97.27$\pm$0.15} \\
			&AutoGEL-intra &\underline{99.85$\pm$0.06}&\underline{97.65$\pm$0.21}&98.92$\pm$0.23&92.36$\pm$1.13&98.29$\pm$0.49&\underline{99.18$\pm$0.09}&97.16$\pm$0.13 \\
			&AutoGEL-diff &99.58$\pm$0.17&97.05$\pm$0.19&98.92$\pm$0.27&90.38$\pm$0.64&97.89$\pm$0.69&98.90$\pm$0.10&96.12$\pm$0.21 \\
			&AutoGEL-$\setminus\delta$ &\underline{99.85$\pm$0.06}&\underline{97.65$\pm$0.20}&\underline{98.98$\pm$0.23}&\underline{92.58$\pm$1.14}&\underline{98.33$\pm$0.39}&99.14$\pm$0.09&\underline{97.23$\pm$0.07}\\
			&AutoGEL-darts &\underline{99.85$\pm$0.06}&97.31$\pm$0.09&98.87$\pm$0.23&91.98$\pm$0.77&97.98$\pm$0.42&99.02$\pm$0.13&95.84$\pm$0.29 \\
			\bottomrule
		\end{tabular}
	}
	\vspace{-10px}
\end{table}

\begin{table}[t]
	\caption{MRR and Hits@N for LP task on knowledge graphs}
	\label{tab:ablation_kg}
	\centering
	\setlength\tabcolsep{5pt}
	\scalebox{0.9}{
		\begin{tabular}{c|c|cccc|cccc}
			\toprule
			\bf Type&\multirow{2}{*}{\bf Model} &\multicolumn{4}{c|}{\bf FB15k-237}&\multicolumn{4}{c}{\bf WN18RR}\\
			&&MRR&Hits@10&Hits@3&Hits@1&MRR&Hits@10&Hits@3&Hits@1\\
			\midrule
			\multirow{2}{*}{Geometric} &TransE 
			&.294&.465&-&-&.226&.501&-&-\\ 
			&RotatE 
			&.338&.533&.375&.241&\underline{.476}&\bf.571&\underline{.492} &.428\\ \midrule
			\multirow{2}{*}{Bilinear} &DisMult 
			&.241&.419&.263&.155&.430&.490&.440&.390\\
			&ComplEx 
			&.247&.428&.275&.158&.440&.510&.460&.410\\ \midrule
			\multirow{4}{*}{NN-based} &ConvKB 
			&.243&.421&.371&.155&.249&.524&.417& .057\\
			&ConvE 
			&.325&.501&.356&.237&.430&.520&.440&.400\\
			&ConvR 
			&.350&.528&.385&.261&.475&.537&.489&\underline{.443} \\
			&HyperER 
			&.341&.520&.376&.252&.465&.522&.477&{.436}\\ \midrule
			\multirow{4}{*}{GLP}  &R-GCN 
			&.248&.417&-&.151&-&-&-&-\\
			&SACN 
			&.350 & \bf{.540} & \underline{.390} &.260 & .470 & .540 & .480& .430 \\
			&VR-GCN 
			&.248&.432&.272&.159&-&-&-&-\\
			&CompGCN 
			&.355&.535&\underline{.390}&.264&\bf .479&.546&\bf .494&\underline{.443} \\ 
			\midrule
			\multirow{3}{*}{AutoGNN} 
			&AutoGEL & \bf .357& \underline{.538} & \bf .391& \bf .266&\bf .479&\underline{.549}& \underline{.492}&\bf .444 \\
			&AutoGEL-$\setminus\lambda$ & .355& .533 & .389& \underline{.265}& .470&.532& .486& .434 \\
			&AutoGEL-darts & \underline{.356} & \underline{.538} & \textbf{.391} & \underline{.265} & .472 & .544 & .485 & .434\\
			&AutoGEL-$\setminus \mathbf{h}_e$ & .355 & .531 & .389 & \underline{.265} & .454 & .540 & .483 & .402 \\
			\bottomrule
		\end{tabular}
	}
	\vspace{-15px}
\end{table}

%

\subsection{Ablation Study}
\label{sec:ablation_study}
Apart from the main experiment results shown in Sec. \ref{sec:4},
we also conduct several ablation studies to investigate the influence of different components in AutoGEL and provide additional experiments in this section. 

\textit{1) Impact of Inter-level Design:}
AutoGEL provides various design dimensions from both intra-level (see Sec.\ref{sssec:intra}) as well as inter-level (see Sec.\ref{sssec:inter}).
To study the effect of the proposed inter-level designs, we set a variant, i.e., AutoGEL-intra, where inter-level design dimensions are removed from the search space, and we only conduct operator search from intra-level. 
As shown in Tab.~\ref{tab:ablation_homo} and Tab.~\ref{tab:ablation_extra}, 
AutoGEL-intra achieves competitive performance compared with manually-designed GNN baselines, which illustrates the powerfulness of AutoGEL's intra-level designs.
But
AutoGEL brings more performance gains over AutoGEL-intra by searching
inter-level operators. 
Note that the number of layers 
$L$
for the LP task on KG is usually 1 (see Appendix~\ref{ssec:case}), thus there are no results of AutoGEL-intra in Tab.~\ref{tab:ablation_kg}.

\textit{2) Impact of Pooling Operator:}
In this paper,
we provide pooling operation candidates $R(\cdot) \in \{sum, mean, max\}$ for the LP task on homogeneous graphs. 
As discussed in Sec. \ref{sssec:pool},
DEGNN~\citep{li2020distance} 
utilizes the difference-pooling as $R(\cdot)$.
Here we set a variant, i.e., AutoGEL-diff, where we remove the proposed  pooling candidates from the search space and fix the difference-pooling  instead. 
As shown in Tab.~\ref{tab:ablation_homo}, the fixed difference-pooling method leads to the significant performance degradation, illustrating the strength of AutoGEL's pooling design.

\textit{3) Impact of Separate Weight Transformation Matrices:}
AutoGEL provide novel linear transformation approaches, i.e., we assign neighborhood-type specific matrices $\mathbf{W}_{\delta(u)}^k$ as special attention mechanism for homogeneous graphs, and edge-aware filters $\mathbf{W}_{\lambda(e)}^k$ to incorporate information from different directions for heterogeneous graphs (see Sec. \ref{sssec:intra}). To study the impact of such designs, we provide two variants, i.e., AutoGEL-$\setminus\delta$ (see Tab.~\ref{tab:ablation_homo} and \ref{tab:ablation_extra} ) and AutoGEL-$\setminus\lambda$ (see Tab.~\ref{tab:ablation_kg}), where $\mathbf{W}_{\delta(u)}^k$ and $\mathbf{W}_{\lambda(e)}^k$ are simply replaced by a single $\mathbf{W}^k$. 
Compared with these two variants, AutoGEL achieves better performance cross different graph tasks and datasets.

\textit{4) Impact of Edge Embedding:}
To show the effectiveness of edge embedding $\mathbf{h}_e$ on the LP task, we set another variant, i.e., AutoGEL-$\setminus \mathbf{h}_e$, by removing $\mathbf{h}_e$ from AutoGEL's MPNN and simply replace $\phi(\mathbf{h}_u^k,\mathbf{h}_e^k)$ with $\mathbf{h}_u^k$ in \eqref{eq:agg_heter}. 
Experiment results are shown in Tab.~\ref{tab:ablation_kg}. 
Performance degradation is observed for the AutoGEL-$\setminus \mathbf{h}_e$, especially on the WN18RR dataset, indicating that $\mathbf{h}_e$ is indeed a critical design.

\textit{5) Impact of Stochastic Differentiable Search Algorithm:}
As discussed in Sec.\ref{ssec:snas}, AutoGEL adopts stochastic differentiable search algorithm in SNAS to perform more effective and efficient architecture search. To show its superiority, we also try the deterministic differentiable search algorithm DARTS for AutoGEL, denoted as AutoGEL-darts. 
Tab.~\ref{tab:ablation_homo}, Tab.~\ref{tab:ablation_kg}, and Tab.~\ref{tab:ablation_extra} empirically show the consistent superior performance of the AutoGEL compared with AutoGEL-darts variant cross node/edge/graph level tasks, indicating the effectiveness of AutoGEL's search algorithm.
Besides, we further show that the search cost of AutoGEL is also lower than its AutoGEL-darts variant (see Tab.~\ref{tab:search_time_vs_darts_nc_gc}, and Tab.~\ref{tab:search_time_vs_darts_lp}).



\begin{table}[t]
	\caption{Average accuracy (\%) for node classification and graph classification}
	\label{tab:ablation_extra}
	\setlength\tabcolsep{2pt}
	\centering
	\scalebox{0.96}{
		\begin{tabular}{c|c|ccc|cccc}
			\toprule
			\multirow{2}{*}{\bf Type} &\multirow{2}{*}{\bf Model} & \multicolumn{3}{|c}{\bf Node Classification} & \multicolumn{4}{|c}{\bf Graph Classification}\\
			&&  Cora &  CiteSeer &  Pubmed &  IMDB-B &  IMDB-M &  MUTAG &  PROTEINS\\
			\midrule
			&PATCHYSAN &-&-&-& 71.00 & 45.20 & 92.60 & 75.90 \\
			&DGCNN &-&-&-& 70.00 & 47.80 & 85.80 & 75.50 \\
			Manual&GCN & 88.11 & 76.66 & 88.58 & 74.00 & 51.90 & 85.60 & 76.00 \\
			GNNs&GraphSAGE & 87.41 & 75.99 & 88.34 & 72.30 & 50.90 & 85.10 & 75.90 \\
			&GAT & 87.19 & 75.18 & 85.73 & -& -& -& -\\
			&GIN & 86.00 & 73.40 & 87.99 & 75.10 & 52.30 & 89.40 & 76.20 \\
			\midrule
			\multirow{7}{*}{AutoGNN}&GraphNAS & 88.40 & 77.62 & 88.96 & -& -& -& -\\
			&SANE & \underline{89.26} & \bf 78.59 & \textbf{90.47} & -& -& -& -\\
			&\cite{you2020design} & 88.50 & 74.90 &- &-& 47.80 &-& 73.90 \\
			\cmidrule{2-9}
			&AutoGEL & \textbf{89.66} & \underline{77.66} & \underline{90.00} & \textbf{81.20} & \textbf{56.80} & \textbf{96.14} & \textbf{82.68}\\
			&AutoGEL-intra & 88.93 & 76.33 & 89.73 & \underline{77.62} & \underline{55.58} & 95.98 & 77.96\\
			&AutoGEL-$\setminus\delta$ & 88.88 & 76.55 & 89.96 & 77.44 & 53.88 & 93.75 & 79.3\\
			&AutoGEL-darts & 89.00 & 77.49 & 89.85 & 76.69 & 47.42 & \underline{96.05} & \underline{80.08}\\
			\bottomrule
		\end{tabular}
	}
	\vspace{-10px}
\end{table}

\begin{table}[!t]
	\centering
	\caption{Search time (clock time in seconds) comparison on the node classification (NC) task and graph classification (GC) task}
	\label{tab:search_time_vs_darts_nc_gc}
	\scalebox{0.92}{
		\begin{tabular}{c|c|c|c|c|c|c|c}
			\toprule
			\multirow{2}{*}{}&\multicolumn{3}{c}{Node Classification} & \multicolumn{4}{|c}{Graph Classification} \\
			\cmidrule{2-4} \cmidrule{5-8} 
			& Cora & CiteSeer & PubMed & IMDB-B &IMDB-M &MUTAG &PROTEINS\\
			\midrule
			AutoGEL & 12 & 16 & 19 & 58 & 90 & 2.4 & 56 \\
			AutoGEL-darts & 15 & 31 & 97 & 122 & 138 & 3.8 & 95\\
			\bottomrule
		\end{tabular}
		}
\end{table}

\begin{table}[!t]
	\centering
	\caption{Search time (clock time in hours) comparison on the LP task}
	\label{tab:search_time_vs_darts_lp}
	\scalebox{0.92}{
		\begin{tabular}{c|c|c|c|c|c|c|c|c|c}
			\toprule
			& NS & Power & Router & C.ele & USAir & Yeast & PB &FB15k-237 & WN18RR\\
			\midrule
			AutoGEL & 0.5 & 2.6 & 3.4 & 1.4 & 1.3 & 4.0 & 14.4 & 18.1 & 13.1\\
			AutoGEL-darts & 1.0 & 2.7 & 3.4 & 1.5 & 1.4 & 6.0 & 14.7 & 18.3 & 13.7\\
			\bottomrule
		\end{tabular}
	}
\end{table}


\subsection{Search Efficiency}
\label{sec:search_efficiency}

As mentioned in Sec.~\ref{sec:4.2}, we found that existing GLP modes generally require more computational resources in practice.
Thus, we try to reduce the search cost in the proposed AutoGEL.
Tab.~\ref{tab:run_time_link} reports
the running time (hours) of AutoGEL and several other representative baselines for the LP task on the homogenous graph (HG) and knowledge graph (KG).

From Tab.~\ref{tab:run_time_link}, we can observe that:
On the LP task on HGs (NA, Power, Router, C.ele, USAir, Yeast, and PB), AutoGEL runs quite fast, which substantially eases the difficulty of using AutoGEL. Besides, AutoGEL achieves more significant performance boost in this scenario (see Tab.~\ref{tab:result_homo}).
On the LP task on KGs (FB15K-237, WN18RR), DistMult \citep{yang2014embedding} is a representative of bilinear models that run much faster among all KGE models. Although AutoGEL is slower than DistMult, its computational cost is very close with classic neural networks (NNs) for KG ConvE \citep{dettmers2018convolutional} and GLP model CompGCN \citep{vashishth2019composition}. Then recalling Tab.~\ref{tab:result_kg}, AutoGEL well balances between search cost and effectiveness.

\begin{table}[!t]
	\centering
	\caption{
		Running time (clock time in hours) of AutoGEL and several baselines for the LP task}
	\label{tab:run_time_link}
	\setlength\tabcolsep{2pt}	
	\scalebox{0.92}{
		\begin{tabular}{c|c|c|c|c|c|c|c|c|c|c}
			\toprule
			\multirow{2}{*}{Type} & \multirow{2}{*}{Model} & \multicolumn{7}{|c}{HGs} & \multicolumn{2}{|c}{KGs} \\
			\cmidrule{3-9} \cmidrule{10-11} 
			& &NS &Power &Router &C.ele &USAir &Yeast &PB & FB15k-237 & WN18RR \\
			\midrule
			GLP for HG & DE-GNN & 0.1 & 1.0 & 1.2 & 0.2 & 0.3 & 2.0 & 4.7 & - & - \\
			Bilinear for KG & DistMult & - & - & - & - & - & - & - & 2.6 & 0.4 \\
			NN for KG & ConvE & - & - & - & - & - & - & - & 26.0 & 10.2 \\
			GLP for KG & CompGCN & - & - & - & - & - & - & - & 16.1 & 7.8 \\
			\midrule
			\multirow{2}{*}{Ours} & AutoGEL (search) & 0.5 & 2.6 & 3.4 & 1.4 & 1.3 & 4.0 & 14.4 & 18.1 & 13.1 \\
			& AutoGEL (training) & 0.3 & 0.5 & 0.7 & 0.4 & 0.4 & 1.5 & 4.8 & 13.1 & 7.3 \\
			\bottomrule
		\end{tabular}
	}
\end{table}



\end{document}